\documentclass[sn-mathphys-num]{sn-jnl}

\usepackage{lmodern}
\usepackage{anyfontsize}
\usepackage{graphicx}%
\usepackage{multirow}%
\usepackage{amsmath,amssymb,amsfonts}%
\usepackage{amsthm}%
\usepackage{mathrsfs}%
\usepackage[title]{appendix}%
\usepackage{xcolor}%
\usepackage{textcomp}%
\usepackage{manyfoot}%
\usepackage{booktabs}%
\usepackage{listings}%

\hypersetup{ hidelinks }
\usepackage{algorithmic}
\usepackage{algorithm2e}
\usepackage{subfigure}
\usepackage{mathtools}
\usepackage{orcidlink}
\newcommand{\readc}[0]{{\em READ-C}}
\newcommand{\readctd}[0]{{\em READ-C-TD}}
\newcommand{\readcsa}[0]{{\em READ-C-SA}}

\begin{document}

\title[Autonomous Curriculum Design via Relative Entropy Based Task Modifications]{Autonomous Curriculum Design via Relative Entropy Based Task Modifications}

\author*[1]{\fnm{Muhammed Yusuf} \sur{Satici}}\email{msatici@ncsu.edu }

\author[1]{\fnm{Jianxun} \sur{Wang}}\email{jwang75@ncsu.edu}

\author[1]{\fnm{David L.} \sur{Roberts}}\email{robertsd@csc.ncsu.edu}

\affil*[1]{\orgdiv{Department of Computer Science}, \orgname{NC State University}, \orgaddress{ \city{Raleigh}, \state{North Carolina}, \country{USA}}}


\abstract{Curriculum learning is a training method in which an agent is first trained on
a curriculum of relatively simple tasks related to a target task in an effort
to shorten the time required to train on the target task. Autonomous curriculum
design involves the design of such curriculum with no reliance on human
knowledge and/or expertise.
Finding an efficient and effective way of autonomously designing curricula
remains an open problem. We propose a novel approach for automatically
designing curricula by leveraging the learner's uncertainty to select curricula
tasks. Our approach measures the uncertainty in the learner's policy
using relative entropy, and guides the agent to states of high
uncertainty to facilitate learning. Our algorithm supports the generation of autonomous curricula in a self-assessed manner by leveraging
the learner's past and current policies but it also allows the use of teacher guided design in an instructive setting. We provide theoretical guarantees for the convergence of our algorithm using two time-scale optimization processes.
Results show that our algorithm outperforms randomly generated curriculum, and
learning directly on the target task as well as the curriculum-learning criteria
existing in literature.
We also
present two additional heuristic distance measures that could be combined with our relative-entropy approach for further performance improvements.}


\maketitle

\section{Introduction}

Curriculum Learning tries to simplify the learning process of an agent on a
difficult-to-learn target task by leveraging the knowledge the agent obtains from learning on a
sequence of source tasks related to the target task. 
Curriculum design for a reinforcement learning problem can be performed either by manually selecting
the tasks or by automated curriculum design algorithms. Automated
curriculum design algorithms are more efficient in choosing tasks as they
eliminate the time spent on manual task generation. On the other hand,
automated curriculum generation requires the crafting of a selection criteria that
can leverage what the agent currently knows and what needs to be learned to converge to the optimal policy faster. By employing autonomous curricula design techniques, the agent aims to attain improved learning efficiency and reduced dependence on human expertise in generation of curricula and achieves this purpose by adaptively modifying the curriculum based on its level of knowledge during training.
We present Relative Entropy based Autonomous Design of Curricula (\readc{}), a
autonomous-curriculum-design framework using the divergence between
policies to identify states with high epistemic uncertainty where additional learning could
improve the performance of the agent. 

Inspiration for \readc{} comes from psychology learning theory.  Studies have
shown that using curriculum-design strategies that reduce students' uncertainty
improves learning efficiency and knowledge acquisition~\citep{adaptUnc,
respondUnc, designUnc, teachingUnc}, and it is often advantageous for students
to successively learn concepts in accordance with some guidance to keep them in
the {\em zone of proximal development}~\citep{Vyg78, Stuyf2002ScaffoldingAA}.
Furthermore, entropy-based sampling approaches have already been used, mostly
in the context of supervised learning and active learning, to query data points
about which the agent has the most uncertainty~\citep{Settles2009ActiveLL}.
We hypothesized that a curriculum-design strategy that reduces
the learner's epistemic uncertainty using the relative entropy (KL divergence) between the agent policy and the true policy
could provide a significant performance improvement to the agent without resulting in a high overhead of curriculum generation. 

\readc{}, at a high level, measures the relative entropy of the regions in the agent's state space
using probability values derived from the agent's policy representation and probability values derived from the true policy, and selects the
states of high uncertainty for the curriculum generation using heuristics based on the relative entropy criteria. It assumes the agent has the ability to change its starting state in the given environment and modifies the Markov decision process (MDP) by moving the start state to a discovered
state of high uncertainty for the generation of each curriculum step. In this setting, relative
entropy encapsulates, for each part of the agent’s state space, how much the learner's current policy diverges
from a true policy standing for an optimal or near-optimal solution for the domain at hand. The higher the
divergence for a region or state of the MDP, the more likely the learner does not have an optimal
policy for that region. \readc{} then modifies the target task to encourage learning in the selected areas of high uncertainty. 

We offer two different realizations of the true policy for
\readc{}, namely \readctd{} for teacher dependent entropy calculation and \readcsa{} for self-assessed entropy calculation.
\readctd{} assumes the existence of a teacher model having learnt the optimal policy and works in a similar fashion
as uncertainty-aware active learning approaches where the agent picks samples for a teacher to label based
on the uncertainty of its learning process w.r.t.\ various entropy metrics including
relative entropy~\citep{SettlesALSequenceLabel}. \readctd{} uses a student-teacher architecture
where the teacher provides the optimal policy for the relative entropy
calculation and the student obtains the uncertainty estimates based on the
relative entropy between the student and teacher policies. 
\readcsa{}, on the other hand, solely relies on the information generated during the agent's training process and estimates the relative
entropy using a regression model with information from the current and past policies of the agent to mitigate the
reliance of the algorithm on already existing policies. Since true policy is often unknown during the training process, using a regressor to estimate the uncertainty values allows the agent to not require the existence of a teacher model for the generation of curricula. The regression model learns from a simpler environment that resembles the target task using the relative entropy between the agent policy and the true policy of the simpler environment as ground truth values in calculation of the uncertainty.

Furthermore,
\readc{} affords different heuristic criteria for prioritizing the regions of uncertainty,
and we present three such criteria: 1) directly using relative entropy, 2)
relative entropy filtered by proximity to a goal state, and 3) relative
entropy filtered by distance to other low-relative-entropy regions. We provide a proof of convergence for the curriculum learning algorithm and explain how the selection of curricula does not change the convergence of the reinforcement learning agent. We evaluate
\readc{} against randomly selected curricula, learning directly on the target
task, and a curriculum-generation criteria from the literature. Results show
that \readc{} outperforms the baseline algorithms and the existing curriculum-selection criteria in many cases, and at worst performs similarly to existing
algorithms while reducing curriculum-generation overhead.

The main contributions of our research are

\begin{itemize}
    \item We present a novel entropy-based curriculum generation algorithm for the purpose of facilitating the learning process of a reinforcement learning agent.
    \item We provide two realizations of this curriculum learning algorithm, a teacher-student framework and a self-assessed approach, to reduce the dependence of the curriculum learning methods on already learnt policies.  
    \item We provide a novel convergence proof for our curriculum generation algorithm to have a theoretical guarantee for the convergence of our reinforcement learning model.
    \item We provide extensive evaluations of our algorithm against curriculum learning and reinforcement learning methods from the literature in three different domains with varying degrees of difficulty. 
\end{itemize}

\section{Related Work}

Curriculum-design techniques are either automated or use a human-in-the-loop
paradigm. Human-designed approaches focus on understanding how humans teach
using curricula~\citep{khanHumans}, and use human knowledge to exploit aspects
of the target task that might improve training efficiency~\citep{PengHuman,
MacAlpineHuman}. Automated approaches generate a sequence of tasks, samples, or
states without human intervention. We can further divide autonomous approaches
into subgroups based on the type of curricula they generate. Some algorithms
provide learners a meaningful sequence of data samples or
demonstrations~\citep{RenPrioritized, seitazpd}, while others generate new
initial/goal/terminal states at every curriculum step~\citep{florensaGoal,
zhangEpistemic}. Other approaches search a parameterized task space by sampling
environment parameters based on the agent's current
knowledge using contextual reinforcement learning~\citep{klink2020selfpaced, portelas2019teacher}. Some further approaches attempt to remove the dependence on the parameterized task distributions by using optimal transport to generate a curriculum in a contextual RL setting \citep{gradientOPT, constOPT}. Some lifelong learning approaches similar to these curriculum learning techniques decompose the task space into subregions based on diverse model primitives and learn subpolicies for each region of the task space in a bottom-up manner to sequentially solve multiple tasks \citep{jaamasLLL}. Task-sequencing
algorithms either heuristically pre-generate a sequence of tasks before
training~\citep{SilvaObject, svetlik} or adaptively generate curricula by
leveraging the learner's knowledge at various points during
training~\citep{QiaoUrbanCar, matiisenTeacherstudent, NarvekarPolicyChange}.
\readc{} uses adaptive modifications of the agent's start state, so our focus
will be on adaptive curriculum-generation algorithms that perform modifications
to the target task.  

Adaptive curriculum-generation algorithms train on the target task and use the
knowledge they obtain to determine which tasks and/or states benefit the
learner. An approach similar to \readc{} is reversed-curriculum
learning that generates a curriculum of start states by
sequentially expanding backwards from the goal state~\citep{florensaGoal}.
In contrast, \readc{} uses heuristic selection of start states based on the
learner's relative uncertainty. Another approach samples start states
proportional to the Euclidean norm of the gradient of a performance measure
defined on the value function, guiding the learner
towards regions of the environment where nearby states have wildly-different
value function estimates~\citep{performMeasure}. It requires learning a start state selection policy
in addition to the RL policy, which increases the training time. Finally,
\citet{NarvekarPolicyChange} present an adaptive task sequencing algorithm
that selects curriculum tasks based on the maximal change in the learner's
policy, guiding the agent towards tasks that are expected to improve the policy
most. This approach requires the learner to
train on the target task and every source task for every curriculum step, which
results in significant curriculum-generation overhead.
See~\citet{narvekar2020curriculum} for more analysis of curriculum learning approaches.

\readctd{} also relates to active learning in that it allows a student to
query a teacher's policy for the purpose of identifying high uncertainty
regions. \citet{SettlesALSequenceLabel} use KL Divergence of an ensemble
of models to identify the samples that the model is least confident in how to
label. \citet{gnnAL} test multiple
uncertainty measures including information gain between the model predictions
and model posterior distribution for graph neural networks.
Our algorithm shows similarities to these approaches in the way it uses relative entropy to determine the uncertainty of the agent regarding its
learning process. However, we adapt this active learning process to a
reinforcement learning problem and measure the agent's uncertainty in its state
space rather than data instances. \citet{jaamasAL} offers an active learning framework using goal-driven demonstrations where the RL agent identifies in which states the feedback is most needed based on Bayesian or quantile confidence and asks for guidance from an expert only for the states it has low confidence of reaching its goal. In contrast, we offer a curriculum learning framework that generates a curriculum for the RL agent without using demonstrations from an expert through a self-assessed learning approach.
\citet{Settles2009ActiveLL} 
explains more about the traditional active learning approaches.

\readc{} relates to options learning literature due to the interleaved learning process it performs based on the uncertainty of the agent over different regions of the state space. Options in the context of options learning represent higher-level actions or sub-policies that the agent learns to execute. By using these options, the agent manages to skip parts of the training process where it already has a good understanding of optimal policy. The traditional options learning algorithms use tabular Q-learning and planning methods to explore the option space \citep{SUTTONOPT, optClas}. Adaptive Skills, Adaptive Partitions framework simultaneously learns options for sub-regions of the state space using a modified gradient calculations based on the hyperplanes that partition the state space into subtasks \citep{mankowitz2016adaptive}. Another sparse sampling approach uses a generative model to sparsely sample a series of states that would construct a small MDP to be used for planning the optimal path~\citep{sparseSampling}. In more recent deep-RL research, actor-critic architectures have been used to model the agent and the options policies \citep{harb2017waiting, bacon2016optioncritic, harutyunyan2019termination}. \citeauthor{bacon2016optioncritic} offers an option-critic architecture where they train the option networks using stochastic gradient descent similar to how an actor would learn and they use a planner to determine which option policy to run at the current step of training ~\citep{bacon2016optioncritic}. \citeauthor{harb2017waiting} improves upon the option-critic by using a deliberation cost model instead of a planner to select good options~\citep{harb2017waiting}. \citeauthor{harutyunyan2019termination} adds a termination critic to the option-critic architecture to optimize the termination condition of the options learning framework ~\citep{harutyunyan2019termination}. Although these options learning algorithms show similarities to \readc{}, their focus lies on skipping the actions that do not offer additional information to the agent whereas the curriculum learning algorithm tries to bring the agent close to the regions of the state space where the learning could improve the policy the most.   

\section{Problem Formulation}

Here we provide background on the RL method and formally define the curriculum-learning problem in this context.

\subsection{Reinforcement Learning}

We model  the learning process as a Markov decision process $M$ $=$ $<S$, $A$, $T$,
  $R$, $s_i$, $S_g, \gamma>$, a tuple consisting of a set of states $S$, a set of actions
$A$, a transition function $T$, a reward function $R$, an initial state $s_i$, a set of terminal states $S_g$ and a discount factor $\gamma$. The transition function $T:S \times A
\times S \rightarrow [0,1]$ corresponds to the probability of transitioning
from a state $s \in S$ to another state $s' \in S$ using a valid action $a \in
A$ at state $s$. All of the environments we use are deterministic, so taking
action $a$ in state $s$ always transitions into the same resulting state
$s'$---although that is not a requirement for \readc{}. The reward function
$R:S \times A \times S \rightarrow F$ maps from a state, action, state tuple to
a real number corresponding to the reward signal the learner receives. The
policy $\pi: S \rightarrow A$ maps states to actions. The cumulative reward $G$
at time $t$ is the discounted sum of all feedback the agent receives from $t$
until it reaches a terminal state, 
\begin{equation*}
    G_t = \sum_{k=0}^{\tau-t} \gamma^{k} R_{t+k}
\end{equation*}
where $\gamma$ is the discount factor, $\tau$ is the time to reach a
terminal state, and $R_{t+k}$ is the feedback received at state
$s_{t+k}$~\citep{SuttonRL}. The agent's objective is to learn the optimal policy
$\pi^*_M$ that maximizes $G$.

For discrete action spaces, we train using dual deep Q-networks (Dual-DQN)~\citep{humanLevelNN}. We use two
four-layered, fully-connected, feed-forward networks. The networks take the
states as input and output Q-value estimates for each available action. One
neural network serves as the learned model, and the other provides target Q-value
estimates for batch updates. The loss function is
\begin{equation}\label{eq1}
    L(\theta) = E_{s, a, r, trm, s' \sim RB} \bigg[ \frac{1}{2} (r + \gamma \text{ max}_{a'} Q(s', a'; \theta^-)  - Q(s, a; \theta))^2 \bigg]
\end{equation}
where $\theta$ is the weights of the learned model, $\theta^-$ denotes the
weights of the target model, $\gamma$ is the discount factor, $s$ is the current state, $a$ is the action taken at state $s$, $s'$ is the next state and $a'$ is the action taken at state $s'$
~\citep{humanLevelNN}. 

For continuous action spaces, we employ an actor-critic architecture similar to~\citep{mnih2016asynchronous}. We use a
four-layered, fully-connected, feed-forward network for the actor and critic. We allocate 256 nodes at every layer and employ rectified linear unit (reLU) activation function between each layer. The critic receives the state of the environment as its input and outputs the value function estimate. The actor network outputs two real vectors which we treat as the mean and standard deviation of the multi-dimensional normal distribution that we sample the actions from. We use the advantage loss to train the actor which is given as 
\begin{equation*}
    L_a(\theta, \omega) = E_{s, a, r, trm, s' \sim RB} \bigg[ log(\pi(a ; s, \theta)) \left( r + \gamma \text{ max}_{a'} V(s'; \omega^-)  - V(s; \omega) \right) \bigg]
\end{equation*}
where \(\omega\) is the learned critic weights, \(\omega^-\) is the
target critic weights, \(\theta\) is the actor weights, $\pi$ is the actor policy that is modeled as a normal distribution and \(\gamma\) is the discount
factor~\citep{mnih2016asynchronous}.
We train the critic network using the mean square error which is given as, 
\begin{equation}\label{eq11}
    L_c(\theta, \omega) = E_{s, a, r, trm, s' \sim RB} \bigg[ \frac{1}{2} (r + \gamma \text{ max}_{a'} V(s'; \omega)  - V(s; \omega))^2 \bigg]
\end{equation}
where $\omega$ is the weights of the critic, $\omega^-$ denotes the
weights of the target critic model, and $\gamma$ is the discount
factor. We use the $epsilon$-greedy policy same as the DQN mentioned above.

At each training step the agent takes a single
action in the environment, records the tuple $(s, a, r, trm, s')$ into a
replay buffer (RB) (where $trm$ indicates whether $s$ is a terminal state),
randomly samples a batch of tuples from the buffer, and performs a single batch
update on the learned model. The target model weights are updated using
the weights of the learned model at the end of each episode.

\subsection{Defining Curricula}

A \readc{} curriculum of length $d$ consists of a sequence of start states $C =
(s_1 \dots s_d)$ such that $s_i \in S$ for all $i \in [1, d]$. Let
$\pi_{i}$ be the agent's policy after training on MDPs $M_1$ to $M_i$ employing
the start states $s_1$ to $s_i$. Then, a curriculum step $i$ produces
the transition $\pi_{i} \rightarrow \pi_{i+1}$ where policy
$\pi_{i}$ is the starting policy for the curriculum step and $\pi_{i+1}$ is the
policy attained by training on MDP $M_{i+1}$. Hence, each curriculum step is a
modification to the agent's policy based on the current MDP $M_{i+1}$. Note
that all of the MDPs in our formulation use the same state space, action space,
dynamics, and reward function.
We compare performance using either the total reward the agent achieves during
training (asymptotic) or the total time spent on the curriculum and target task
to obtain the maximum cumulative reward $G^{max}$ (time to convergence).

The objective of the curriculum learning algorithm is to improve the sample efficiency and the learning speed of the reinforcement learning process. In environments where it is often costly to retrieve samples, it is advantageous to design guidance methods that could direct the agent towards the parts of the state space that the training would create improvements in the learnt policy of the agent. \readc{} aims to modify the start state of the environment with the purpose of bringing the agent closer to the states it is struggling to learn to facilitate the learning process and reduce the number of samples needed from the environment to reach the optimal policy. Since all of our algorithms perform one batch update after receiving each sample from the environment, reducing the sample complexity, in this case, also reduces the number of batch updates made during the learning process, speeding up the training time of the agent.   

\subsection{Transfer Learning}\label{transfer}

The learner retains the same model throughout training, using the model
obtained in the prior curriculum step as the starting point for the next
step. Generating new models or transferring weights is not
required. The learner also retains the contents of the replay buffer across
curriculum steps.

\section{Curriculum Learning}

We first describe our uncertainty measures and provide pseudo-code for the high-level description of \readc{}. Then, we define two main realizations of \readc{}, namely \readcsa{} and \readctd{}, and describe two distance-metric variants of the relative-entropy heuristic. 

\subsection{Measuring Uncertainty}

We measure the uncertainty between two policies, namely the learnt policy and the true policy, through relative entropy. The relative entropy is defined as
\begin{equation}\label{eq6}
     D_{KL}(P_{true} || P_{learnt}) = 
     \sum_{a \in A} P_{true}(s, a)log\Bigg(\frac{P_{true}(s, a)} {P_{learnt}(s, a)}\Bigg)
\end{equation}
where $P_{true}$ denotes the action probabilities obtained from a reference policy and $P_{learnt}$ represents the probabilities of
a learnt policy at a certain checkpoint during the training. We take the relative entropy of the reference probabilities with
respect to the learnt probabilities to measure the reduction in
uncertainty the agent would have in its policy if it were to use the
correct policy instead of its own. In this manner, our uncertainty metric reflects how inaccurate the agent is in its action selection for a given state of MDP. Since the true policy already has good performance on the target task, the relative entropy is expected to drop as the agent improves its policy and comes closer to the estimate of the true policy. The probability distribution here could be defined over a discrete or continuous variable depending on how the neural networks treat the output parameters.

Since the policy relies on the Q-value estimates of the agent, we require a transfer function that could convert the Q-value estimates to probability values for the relative-entropy calculation. For this purpose, we use softmax and calculate the probability of action selection as

\begin{equation*}
    P(s, a) = \frac{e^{\frac{Q(s, a)} {||Q(s)||_2}}} {\sum_{a \in A} e^{\frac{Q(s, a)} {||Q(s)||_2}}}
\end{equation*}
where $Q(s,a)$ is the Q-value for the state, action pair $(a,s)$, $Q(s)$ is the
list of Q-values for state $s$.

\subsection{\readc{}}

Given the definition of uncertainty in Eq. \ref{eq6}, we construct a high-level curriculum-learning algorithm that affords the use of different RL algorithms for the training of the agent. Algorithm~\ref{alg:algo1} is the high-level pseudo-code for \readc{}. It initializes the target environment and the agent model (Line~4). It starts with an empty curriculum (Line~5) and trains the agent model for $\eta$ iterations using the train function defined in Algorithm~\ref{alg:algo2} to generate
an initial agent policy (Line~7)--- the starting point for relative-entropy calculation. Each iteration of the loop (Line~8) creates a new curriculum step by selecting a new start state and the agent trains on MDP with the new start state. The function named uncertainty defined in Algorithm~\ref{alg:algo3} (Line~9) uses Eq. \ref{eq6}  to calculate relative entropy values of current policy for each state in the set of visited states ($SB$), and returns the state with the highest uncertainty to Algorithm~\ref{alg:algo1}. Then, Algorithm~\ref{alg:algo1} changes the start state to the state of highest uncertainty (Line~10) and trains on the modified MDP until the convergence criteria---if the entropy has not
reduced for the last 10 episodes---is met (Line~12). This convergence method receives the relative entropy of the selected state from the agent before and after each training episode, which is not computationally expensive since the algorithm calculates entropy for only a single state---not all regions. The process repeats until the curriculum reaches a defined length, at which
point the student trains on the original target task until an overall
convergence criterion---if the agent has not
reached the highest cumulative reward for the last 10 episodes---is met (Line~17). 

\begin{algorithm}[htb!]
    \caption{\readc{}}
    \label{alg:algo1}
\begin{algorithmic}[1]
\small
    \STATE \textbf{Inputs:} target MDP: $M_{tar}$.
    
    \STATE \textbf{Consts:} \# of training iterations: $\eta$; curriculum length threshold: $MAX\_LENGTH$.

    \STATE \textbf{Vars:} learning model: $NN$; environment: $ENV$; curriculum: $C$; state buffer: $SB$; convergence criteria: $conv$. 

    \STATE $NN_{agent}, ENV_{tar} \gets initialize(M_{tar})$

    \STATE $C \gets \{\}$

    \STATE Set $conv$  to $\eta$ training iterations

    \STATE train($NN_{agent}$, $ENV_{tar}$, $conv$, $SB$)
    
    \WHILE{$|C|$ $<$ $MAX\_LENGTH$}
    \STATE $s \gets uncertainty(NN_{agent}, ENV_{tar}, SB)$
    
    \STATE $ENV_{cur} \gets$ Set $s$ as the start state of $ENV_{tar}$ 
    
    \STATE Set $conv$ to 10 episodes of no entropy reduction
    
    \STATE train($NN_{agent}$, $ENV_{cur}$, $conv$, $SB$)
    
    \STATE $C \gets C \cup s_i$
    \ENDWHILE 

    \STATE Set $conv$ to highest cumulative reward 
    
    \STATE Make $s_i$ of original $M_{tar}$ the $s_i$ of $ENV_{tar}$
    
    \STATE train($NN_{agent}$, $ENV_{tar}$, $conv$, \_)
    
\end{algorithmic}
\end{algorithm}

\begin{algorithm}[htb!]
\caption{Train}
\label{alg:algo2}
\begin{algorithmic}[1]
\small
\STATE \textbf{Inputs:} learning model: $NN_{agent}$; environment: $ENV$; convergence criteria $conv$, state buffer: $SB$.

\STATE \textbf{Consts:} set of terminal states of $ENV$: $S_g$; initial state of $ENV$: $s_i$.

\STATE \textbf{Vars:} replay buffer $RB$. 

\WHILE{$conv$ is not satisfied}

    \STATE $s \gets s_i$ of $ENV$ 
    
    \STATE $trm \gets s \in S_g$    
    
    \WHILE{$trm$ is False}
    
    \STATE Select $a$ using $\epsilon$-greedy policy on $NN_{agent}(s)$
    
    \STATE Take action $a$ in $ENV$ and observe $r, s'$
    
    \STATE Store $(s,a,r,trm,s')$ in RB
    
    \IF{$s \notin SB$}
        \STATE Store s in $SB$
    \ENDIF
    
    \STATE Sample a minibatch B from $RB$
    \STATE Perform an optimization step on $NN_{agent}$ using $B$
   
   \STATE $s \gets s'$
    \STATE $trm \gets s \in S_g$   
    \ENDWHILE    
\ENDWHILE
\end{algorithmic}
\end{algorithm}

Algorithm~\ref{alg:algo2} describes the train function used in Algorithm~\ref{alg:algo1}. It receives the learning model, the current environment, the convergence criteria and the state buffer as inputs and performs regular RL training on the environment using the learning model until the convergence criteria is met (Line~4). It initializes the environment at the start state (Line~5) and selects the action to take using an $\epsilon$-greedy policy on the learning model (Line~8). Then, it takes the selected action on the environment (Line~9) and stores the (s,a,r,trm,s') tuples in $RB$ for future training (Line~10). It also stores every unique state visited so far in $SB$ for the entropy calculation (Line~12). Then, it performs a single step of optimization on the learning model using a minibatch of samples from $RB$ (Line~15). Algorithm~\ref{alg:algo2} affords the use of different learning models and policies as it (neither \readc{}) does not rely on any specific policy or network architecture for the entropy calculation or agent training. It can also be easily adapted to non-episodic environments by removing $trm$ variable (Line~6) and replacing the loop that checks for terminal states (Line~7) with the outer loop that checks a suitable $conv$ for non-episodic training (Line~4).

\subsection{Relative-Entropy Calculation for \readctd{}}\label{teach}

\begin{algorithm}[tb]
\caption{Relative-Entropy Calculation for READ-C-TD}
\label{alg:algo4}
\begin{algorithmic}[1]
\small
\STATE \textbf{Inputs:} learning model: $NN_{agent}$; environment: $ENV$; state buffer: $SB$.

\STATE \textbf{Consts:} a teacher model: $T$.

\STATE \textbf{Vars:} Q-values for state s: $Q^s$; probabilities for state s: $P^s$. 

\STATE \textbf{Outputs:} the new start state: $s$.

\STATE states $\gets$ Sample a subset of states from $SB$
 
\STATE $UNCERTAINTIES[0, ..., N] \gets 0$ where N is the number of states

\FOR{$s_j$ in $states$}
    
   \STATE $Q_{agent}^s \gets$ get Q-values of $s_j$ from $NN_{agent}$
    
   \STATE $Q_{teach}^s \gets$ get Q-values of $s_j$ from  $T$
    
   \STATE $P_{agent}^s \gets softmax(normalize(Q_{agent}^s))$
    
   \STATE $P_{teach}^s \gets softmax(normalize(Q_{teach}^s))$
    
    \STATE $UNCERTAINTIES[j] \gets$ $\mathrlap{entropy(P_{teach}^s, P_{agent}^s)}$
    
\ENDFOR
\STATE $k \gets$ argmax of $UNCERTAINTIES$
\STATE return $s_k$

\end{algorithmic}
\end{algorithm}

Algorithm~\ref{alg:algo4} describes the relative-entropy calculation for the teacher dependent variant of \readc{}. It necessitates the use of a teacher model that already knows the optimal policy of the target environment to compute the relative entropy values for curriculum selection. If there happens to be an already trained model for the target MDP that could serve as the teacher, Algorithm~\ref{alg:algo4} offers a simpler way of calculating the uncertainty metric as it does not require $\beta$ iterations of training Algorithm~\ref{alg:algo3} does on the learning model to generate two policies of the agent at different checkpoints. On the other hand, if the teacher needs to be trained from scratch, Algorithm~\ref{alg:algo3} offers a much faster performing framework. If there exist an agent residing in a remote server which the user could query to measure the relative entropy but cannot copy its weights directly to the agent due to privacy and/or security reasons, then that agent could be used in \readctd{} as a teacher model, which present one possible use case of \readctd{} despite its requirement of already having a learnt model. A more concrete example of this situation would be commercial LLMs where we are not given open source access to the neural network itself but we can interact with the model through an API. In a similar sense, when a deep RL model is used in a copyrighted setting, we can interact with the model to generate action probabilities without having the need to access the inner mechanisms of the model. Then, using the output of the model allows me to perform training on \readctd{} and get the benefit of teacher dependent curriculum generation.

Algorithm~\ref{alg:algo4} contains a teacher model that is trained to convergence using Algorithm~\ref{alg:algo2} without employing any curriculum learning. Algorithm~\ref{alg:algo4} receives the agent model, the environment and the state buffer $SB$ as its inputs and uses the teacher along with the information it has on the agent to calculate relative entropy. Algorithm~\ref{alg:algo4} calculates the probability values based on the Q-values estimates of the sampled states (Lines~10~\&~11). Then, it calculates the relative entropy between the agent's and teacher's policies and uses it as the uncertainty value for the given state (Line~12). Since the teacher knows the optimal policy, the relative entropy, in this case, shows how much the agent's knowledge is diverging from the optimal policy. Finally, \readctd{} chooses the highest-uncertainty state as the new start state (Line~14).

\subsection{Relative-Entropy Calculation for \readcsa{}}\label{noteach}

Algorithm~\ref{alg:algo3} describes the $uncertainty$ function given in Algorithm~\ref{alg:algo1} at Line~9 and estimates the relative entropy for the self-assessed version of \readc{}. \readcsa{} uses a regression model to estimate the relative entropy between the agent and teacher and gets rid of the requirement of having a teacher model already trained in the target environment. The regressor serves as a proxy function for estimating the true relative entropy using the information theoretic data generated on a simpler environment. It attempts to learn an underlying relationship between the agent's knowledge of the task and the agent's uncertainty of the action selection. Since \readcsa{} uses the true policy of the simpler environment in its training, the true policy in this setting does not necessarily represent how the agent should act in the target environment but due to the simpler environment being a subtask of the target environment having similar characteristics, it is expected that the knowledge acquired on the simpler task would be generalizable to the target task in calculation of the uncertainty. \readcsa{} offers a compromise in that it allows us to predict the relative entropy values more efficiently without relying on the teacher policy but it also introduces additional error coming from the training of the regression model, which we try to minimize by selecting a proper regression model for the problem at hand.  

\begin{algorithm}[tb]
\caption{Relative-Entropy Calculation for READ-C-SA}
\label{alg:algo3}
\begin{algorithmic}[1]
\small
\STATE \textbf{Inputs:} learning model: $NN_{agent}$; environment: $ENV$; state buffer: $SB$.

\STATE \textbf{Consts:} \# of training iterations: $\beta$.

\STATE \textbf{Vars:} Q-values for state s: $Q^s$; probabilities for state s: $P^s$; convergence criteria: $conv$. 

\STATE \textbf{Outputs:} the new start state: $s$.

\STATE states $\gets$ Sample a subset of states from $SB$
 
\STATE $UNCS[0, ..., N] \gets 0$ where N is the number of states

\STATE Set $conv$  to $\beta$ training iterations

\STATE $NN_{past} \gets NN_{agent}$

\STATE train($NN_{agent}$, $ENV$, $conv$, $SB$)

\FOR{$s_j$ in $states$}
    
   \STATE $Q_{past}^s \gets$ get Q-values of $s_j$ from $NN_{past}$
    
   \STATE $Q_{agent}^s \gets$ get Q-values of $s_j$ from  $NN_{agent}$
    
   \STATE $P_{past}^s \gets softmax(normalize(Q_{past}^s))$
    
   \STATE $P_{agent}^s \gets softmax(normalize(Q_{agent}^s))$
    
   \STATE $relEntropy \gets$
   $\mathrlap{entropy(P_{agent}^s, P_{past}^s)}$
    
    \STATE $entropyCur \gets$
   $\mathrlap{entropy(P_{agent}^s)}$
   
    \STATE $entropyPast \gets$
   $\mathrlap{entropy(P_{past}^s)}$
    
    \STATE $UNCS[j] \gets reg(relEntropy, entropyCur,$  $entropyPast,  Q_{past}, Q_{agent})$
    
\ENDFOR
\STATE $k \gets$ argmax of $UNCS$
\STATE return $s_k$

\end{algorithmic}
\end{algorithm}

Algorithm~\ref{alg:algo3} samples a subset of the states in $SB$ for the uncertainty calculation (Line~5). This step is crucial in ensuring that the algorithm could run in large state spaces especially if the state space is continuous. Then, the algorithm performs training for $\beta$ iterations to generate a second learning model (Line~9). The two learning models of the agent, $NN_{past}$ and $NN_{agent}$, at two different checkpoints in time helps us measure how much the agent is diverging from its own policy as time passes. Although the relative entropy calculated between $NN_{past}$ and $NN_{agent}$ does not reflect the true relative entropy, it represents the speed of learning and change in the Q-value estimates of the model. For every sampled state, \readcsa{} calculates the Q-value estimates of both models (Lines~11~\&~12) and converts them to probabilities by normalizing and softmaxing (Lines~13~\&~14). 
The algorithm normalizes the Q-values using the L2-norm to make the
results independent of the scale of the Q-values and uses softmax to obtain
probability estimates. It calculates relative entropy and entropies on the probability values using Eq.\ref{eq6} and \ref{eq7} (Lines~15,~16~\&~17). By doing these operations, \readcsa{} generates information theoretic data about the agent that we use for estimating the relative entropy for the given state. We pass the data to a regressor model that was trained on the relative-entropy values of a smaller and easier-to-train environment. Since the input features to the regressor are independent from the state and action dimensions of the environments, we can deploy the trained regression model from a simpler environment to the target environment without any modification. The regressor model serves as a proxy function for the relative entropy between the true policy and the agent policy of the target environment as long as the easier-to-train environment shares the same characteristics as the original task. The regressor predicts the uncertainty of the states based on the relative entropy, entropy and q-value estimates of the agent policy and records it in an array (Line~18). Finally, the algorithm chooses the maximum uncertainty state as the new start state (Line~20) and returns it (Line~21).  

Our regressor is a gradient-boosting machine (GBM). It trains on the data (relative entropy, entropy, Q-values and number of visits per state) generated by a learning model of a simpler environment. In addition to relative entropy, we also use the entropy for each policy in the training of the regression model. The entropy is defined as
\begin{equation}\label{eq7}
    H(\pi(\cdot | s)) = - \sum_{a \in A} P(s, a)log(P(s, a)),
\end{equation}
where \(H(\pi(\cdot | s))\) is the entropy of the action policy $\pi$ for the state \(s\) using the probabilities \(P\) coming from a given \readc{} learner's policy. The training of the regressor follows the same pattern as Algorithm~\ref{alg:algo3} to generate the data (Lines~13,~14,~15,~16~\&~17) but instead of using the data to predict the uncertainty values, the regressor learns a mapping from the data to the relative entropy between the agent and the optimal policy. As long as the trained environment shows a high level of similarity to the target MDP, the regressor generates accurate uncertainty estimates for every sampled state. Since we employ a simpler MDP for the regressor optimization, the training of GBM takes much less time than the curriculum-learning process allowing READ-C-SA to efficiently calculate uncertainty values without having a need to measure the true policy on the target MDP.

The regression model takes the information theoretic data (relative entropy, entropy, Q-values and number of visits per state) generated on the simplified version of the target task as input and uses the true relative entropy as the ground truth value for its training. The true relative entropy is calculated between the agent policy and the policy of a teacher model trained on the simpler environment. Since it is easier to train on the simplified environment than the target task, generating the teacher model in the case of \readcsa{} takes much less time than \readctd{}, allowing the curriculum generation process to be more efficient. The regression model uses the mean squared error between its output and the true relative entropy as the loss function. Once the regression model trains on the data generated for every visited state of the simpler environment, it gets deployed in the target environment to predict the uncertainty values, which is shown in Algorithm \ref{alg:algo3}. We train different types of regressors, namely Linear Regression, Random Forest Regression, Support Vector Machine and GBM, and choose the best performing regressor (GBM) to be deployed in the target environment.

\subsection{Clustering of States}

Optionally, one could also perform an Agglomerative Hierarchical
Clustering~\citep{scikit-learn} over sampled states, merging clusters using the Ward criterion~\citep{Ward63} until a distance cutoff is reached, and calculate the uncertainty values over the clusters of states rather than single states. In this case, the fundamental flow of the algorithm remains unchanged as it still calculates a single uncertainty value for every state but the uncertainty values of the states in the same cluster gets aggregated at the end of the algorithm to find the uncertainty of each cluster. Then, the algorithm chooses the highest-uncertainty cluster and selects a random state from that cluster as the new start state. The clustering of the states enables us to reduce the deviation in our measurements since single states might show high variance in their uncertainty from one run of the algorithm to the other. It is more advisable to perform clustering if the nature of the state space allows such a partitioning but in domains where the state space cannot generate meaningful clusters, the algorithm needs to run in its original form without clustering. 

\subsection{Heuristic Variants}\label{var}

\readc{} as described above chooses the maximum relative entropy, which does
not account for the underlying structure of the problem. For example, a region
of high uncertainty may be located in an area of the state space extremely far
from the start and goal states of the target task, or that may be rarely
encountered when following an optimal policy. Relative entropy alone would
prioritize training in such regions even if not advantageous. Therefore, we
present two additional relative-entropy heuristics: 

1) Proximity: proximity to {\em positive reward terminal states}, which filters out 80\%
of the regions having the worst proximity (average Euclidean distance in our
case) to terminal states that provide a positive reward, and computes relative
entry only for the remaining 20\%; 

2) Max Distance: proximity to {\em low-entropy
regions}, which classifies regions as high-entropy (one standard deviation above the mean relative entropy) and low-entropy (one standard deviation below the mean relative entropy), and sorts
the regions of high-entropy based on the (Euclidean, in our case) distance to
the closest low-entropy region (selecting the region with the highest
distance). This technique promotes selected regions not resulting in
lengthy training episodes to improve efficient training.

\begin{figure}[hbt]
    \centering
    \includegraphics[width=0.7\columnwidth]{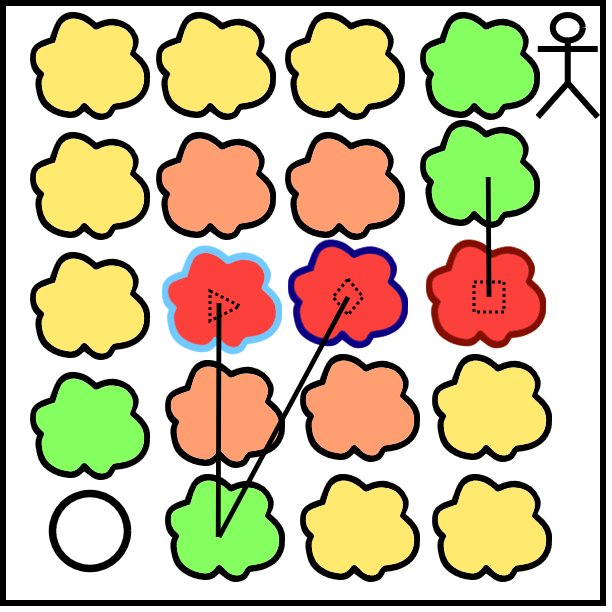}
    \caption{Visualization of the highest uncertainty region for \readc{} variants in an environment with a single goal.}
    \label{fig:fig0}
\end{figure}

Figure~\ref{fig:fig0} depicts a 10 by 10 environment containing a single agent
(straw man) at the top right corner and a positive reinforcement item (white
circle) at the bottom left corner. The rest of the environment is clustered
into regions with varying relative entropies. Green denotes low-entropy
regions, red (containing dotted shapes inside) denotes high-entropy regions,
and yellow and orange denote moderate-entropy regions. Because there are no
actions leading into walls, it is faster to learn the regions close to the edges of the environment; hence, those
regions are mostly low-to-moderate entropy. Similarly,
the regions near the white circle have low entropy because of the short-horizon
goal-state reward. The regions near the center of the environment have
high entropy. Black lines indicate the closest low-entropy region for the high entropy regions. In this situation, relative entropy by itself would prioritize one of
the three red regions, proximity would prioritize the red region with light
blue edges (containing a dotted triangle),
and maximum distance to the low-entropy regions would prioritize the red region
with dark blue edges (containing the dotted diamond).
Therefore, the proximity variant of \readc{} will aim to keep the agent close to the
positive reinforcing terminal state whereas the maximum distance to low entropy variant will aim to
keep it away from the already learned regions. The relative entropy by itself could still choose a suitable region to train on in this example but due to the randomness of the training process caused by the $\epsilon$-greediness, there will be no guarantee that the region of highest uncertainty the relative entropy generates would be the most useful region for the agent to reduce its uncertainty. However, by using the distance metrics, the agent would at least make sure that the region being selected is not far away from the goal states and/or states with useful information regarding the optimal policy path.

\subsection{Proof of Convergence for \readc{}}

Let $L_a$ be the loss function for the actor and $L_c$ be the loss function for the critic. 

\begin{equation}
    L_a(\theta_t, \omega_t, z_t) = E_{\tau \sim \pi(\theta_t, z_t | s_0)} \bigg[ log( \pi(a_t | s_t, \theta_t)) (R_t + \gamma V(s_{t+1} | \omega_t) - V(s_t | \omega_t)) \bigg]
\end{equation}
where $\pi(a_t | s_t, \theta_t) \sim \mathcal{N}(\mu_t,\,\sigma_t^{2})$, $s_t$ is the environment state, $a_t$ is the action taken at time t, $R_t$ is the reward received from the environment, $\gamma$ is the discount factor, $\omega_t$ is the weights of the critic network, $\theta_t$ is the weights of the actor network, V is the value function estimated by the critic, $\pi$ is the policy of the actor network, $z_t$ is the controlled Markov process modeling non-additive noise, $\tau$ is the trajectory of the agent following policy $\pi$ and $(\mu, \sigma)$ pair is the mean and standard deviation of the Gaussian distribution modeling the actor network policy $\pi$.

\begin{equation}
    L_c(\theta_t, \omega_t, z_t) = E_{\tau \sim \pi(\theta_t, z_t | s_0)} \bigg[ \frac{1}{2} (R_t + \gamma V(s_{t+1} | \omega_t) - V(s_t | \omega_t))^2 \bigg]
\end{equation}

The gradient update rule for the loss in advantage actor-critic model is

\begin{equation}\label{eq1n}
    \theta_{t+1} = \theta_t - \alpha(t) (\nabla L_a (\theta_t, \omega_t, z_t) + m_t ) 
\end{equation}
where $\alpha(t)$ is the learning rate for the actor at time t and $m_t$ is the martingale sequence noise.

\begin{equation}\label{eq2}
    \omega_{t+1} = \omega_t - \beta(t) (\nabla L_c (\theta_t, \omega_t, z_t) + m_t ) 
\end{equation}
where $\beta(t)$ is the learning rate of the critic network and $m_t$ is the martingale sequence noise.

\textbf{Theorem 1} Under the assumptions of Section \ref{assm}, Eq. \ref{eq1n} \& \ref{eq2} converges to
\begin{equation}
    (\theta_t, \omega_t) \rightarrow (\theta^*, \omega^*) \textnormal{  as  } t \rightarrow \infty
\end{equation} 

\subsubsection{Assumptions}\label{assm}

In this section, we describe the assumptions needed to prove the convergence of \readc{} to an optima. We use the problem formulation from \citep{karmakar2017timescale} for crafting the assumptions of our proof. We treat the actor-critic learning process as a two time-scale optimization process and make use of the actor-critic convergence proof in proving that the selection of curricula does not change the convergence of the reinforcement learning algorithm.

\textbf{Assumption 1}

$z_t$ is a Markov process taking values in a compact metric space w.r.t. a continuous transition function. 

\textbf{Assumption 2} 
$L_c$ and $L_a$ are J-Lipschitz and J-Smooth functions such that
\begin{equation}
    || \nabla L_a(\theta_t, \omega_t, z_t) || \leq J \quad \forall \, \theta \, \& \, \omega \in {\rm I\!R}
\end{equation}

\begin{equation}
    || \nabla^2 L_a(\theta_t, \omega_t, z_t) || \leq J \quad \forall \, \theta \, \& \, \omega \in {\rm I\!R}
\end{equation}
where $J$ is the Lipschitz constant.

\textbf{Assumption 3}

$m_t$ is a martingale difference sequence with bounded second moment such that

\begin{equation}
    \begin{split}
    E[m_t^2 | F_t] \leq K + \theta_t^2 \\
    E[m_t^2 | F_t] \leq K + \omega_t^2
    \end{split}
\end{equation}
where $F_t$ is history of martingale variables up until time t and $K$ is the martingale constant. 

\textbf{Assumption 4}

\begin{equation}
    \sum_t \alpha(t) = \beta(t) = \infty \, , \, 
    \sum_t (\alpha(t)^2 + \beta(t)^2) < \infty
\end{equation}
for $\alpha(t), \beta(t), t \geq 0$.

\textbf{Assumption 5}

Assume that $z_t$ has an ergodic occupation measure $\Gamma(\theta, \omega)$. Then, let $l_c$ be $\nabla L_c(\theta_t, \omega_t, z_t)$ and $l_a$ be $\nabla L_a(\theta_t, \omega_t, z_t)$. 

\begin{equation}
    \bar{l_c} = \int l_c(\theta, \omega, z) \Gamma(\theta, \omega)   
\end{equation}
where $\bar{l_c}$ is the ordinary differential equations for the gradient of the loss function. 

$\bar{l_c}$ has a global attractor set $B_{\theta}$ and a stable point $\lambda(\theta)$ such that $\lambda(\theta)$ is a J-Lipschitz map.

Furthermore,
\begin{equation}
    \bar{l_a} = \int l_a(\theta, \lambda(\theta), z) \Gamma(\theta, \lambda(\theta))   
\end{equation}
$\bar{l_a}$ has a global attractor set $A$.

\textbf{Assumption 6}
The weights of the neural networks have a tight upper bound such that

\begin{equation}
    sup( || \theta_t || + || \omega_t || ) < \infty
\end{equation}

\subsubsection{Effect of Curriculum Selection on Convergence}

The actor-critic architecture is expected to converge to an optima based on the convergence proofs provided in Wu et al. and Holzleitner et al. \citep{wu2022finite, holzleitner2020convergence} using the assumptions from two time scale optimization processes \citep{karmakar2017timescale}. Under these conditions, we must show that the selection of curriculum states does not violate the convergence of the actor-critic model for \readc{} to also converge.

\readc{ } determines the start state for the environment based on the KL divergence between the agent's policy and the true policy of the MDP which is given as

\begin{equation}\label{kl}
    s_0 = argmax_{s \in S} KL(s | \theta_t, z_t)
\end{equation}

If we assume that the output of the actor-critic model is normally distributed (which is an assumption that holds for advantage actor-critic), then KL divergence will have a closed form solution which is

\begin{equation}\label{kl2}
    KL(s | \theta_t, z_t) = log ( \frac{\sigma_{lrn}(s | \theta_t, z_t)}{\sigma_{true}} ) + \frac{\sigma_{true}^2}{2\sigma_{lrn}^2(s | \theta_t, z_t)} + \\ \frac{(\mu_{true} - \mu_{lrn}(s | \theta_t, z_t))^2}{2\sigma_{lrn}^2(s | \theta_t, z_t)}
\end{equation}
where $\mu$ and $\sigma$ refer to the mean and standard deviation of the Gaussian distribution the policy samples the actions from. $(\mu_{true}, \sigma_{true})$ comes from an already existing optimal policy and $(\mu_{lrn}, \sigma_{lrn})$ comes from the current policy of the agent.

The expected loss for following a trajectory initiated at a starting state generated by the curriculum learning algorithm would be
\begin{equation}
    \sum_{s_c \in S} P(s_0 = s_c | KL(s_c | \theta_0, z_0)) P(\tau | s_0, \theta_0) E[L_a(\theta_t, \omega_t, z_t, s_t)]
\end{equation}
where CL refers to the set of start states $s_0$ calculated by Eq.\ref{kl} w.r.t. KL divergence given in Eq.\ref{kl2}.

Given the convergence of the actor-critic, we wish to show that the distribution of start states above does not change the convergence of \readc{}.

For the first base case, if $s_c = s_g$, then it is trivial to show that the algorithm will converge since the agent is already at a goal state and does not need to take any action to reach convergence.

For the second base case, if $s_c = neighbor(s_g)$, then the expected loss would be
\begin{equation}
    P(s_g | s_0, \theta_0) E[L_a(\theta_t, \omega_t, z_t, s_t)]
\end{equation}
for a sufficiently large exploration rate of $\epsilon$, $P(s_g | s_0, \theta_0)  > 0$ and we know that $E[L_a(\theta_t, \omega_t, z_t, s_t)]$ converges when $t \rightarrow \infty$ based on the actor-critic convergence proof \citep{karmakar2017timescale, holzleitner2020convergence}. Then, the algorithm is supposed to converge to an optima when the start state is a neighbor of a goal state.

Having established the base cases, we now assume that \readc{} converges when $s_c = s_{n}$ where $s_n$ is a state on trajectory $\tau$. We need to show that \readc{} converges when $s_c = s_{n-1}$ where $s_{n-1}$ is a prior state to $s_n$ on the trajectory. The expected loss is
\begin{equation}
    P(s_0 = s_{n-1} | KL(s_{n-1} | \theta_{n-1}, z_0)) P(\tau | s_{n-1}, \theta_{n-1}) E[L_a(\theta_{t-1}, \omega_{t-1}, z_{t-1}, s_{t-1})]
\end{equation}
which is equivalent to
\begin{equation}\label{eqCur}
    P(s_0 = s_{n-1} | KL(s_{n-1} | \theta_{n-1}, z_0)) P(s_n | s_{n-1}, \theta_{n-1}) P(\tau | s_{n}, \theta_{n}) E[L_a(\theta_{t-1}, \omega_{t-1}, z_{t-1}, s_{t-1})]
\end{equation}

\begin{equation}
E[L_a(\theta_{t-1}, \omega_{t-1}, z_{t-1}, s_{t-1})] = E[L_a(\theta_{t}, \omega_{t}, z_{t}, s_{t})]-\langle l_a(\Upsilon_{t-1}), \theta_{t} - \theta_{t-1} \rangle
\end{equation}
based on Taylor's Theorem \citep{convexOpt}. As $t \rightarrow \infty$, we know that $|\theta_{t} - \theta_{t-1}| \rightarrow 0$ due to the convergence of the actor, which makes the inner product term 0, leaving us with $E[L_a(\theta_{t}, \omega_{t}, z_{t}, s_{t})]$. Similarly, 

\begin{equation}
 P(s_0 = s_{n-1} | KL(s_{n-1} | \theta_{n-1}, z_0))P(s_n | s_{n-1}, \theta_{n-1}) \propto P(s_0 = s_{n} | KL(s_{n} | \theta_{n}, z_0))   
\end{equation}

Then, we could rewrite Eq.\ref{eqCur} as
\begin{equation}
    P(s_0 = s_{n} | KL(s_{n} | \theta_{n}, z_0))  P(\tau | s_{n}, \theta_{n}) E[L_a(\theta_{t}, \omega_{t}, z_{t}, s_{t})]
\end{equation}
We know that this equation converges due to the inductive assumption. Then, \readc{} must converge for $s_0 = s_{n-1}$.

We have shown that the curriculum selection does not change the convergence of \readc{}. Given that the curriculum selection criteria does not change the convergence, \readc{} must converge under the assumptions given in Section \ref{assm}.

\section{Evaluation Methods}

Here we describe the test environments, the experimental setup, and the state
representation used for each domain.

\subsection{Test Environments}\label{test}

\subsubsection{Key-Lock}

We use a $20\times20$ 2D grid environment as well as continuous state-action space domains to test the algorithms. 
The first domain
contains keys, locks, pits, and obstacles similar to~\citet{KonidarisDomain, NarvekarPolicyChange}.  The agent's task is to pick up the key and
unlock the lock while avoiding the pits and obstacles. Each key picked up gives
a reward of 500 and each lock unlocked gives a reward of 1,000. Falling into a
pit receives -400.  All other actions including moving into an obstacle receive
-10.  Moving into an obstacle results in no state transition. The learner can
only move in cardinal directions and is assumed to have obtained the key or the
lock if its location matches the location of the key or the lock.  An episode
terminates when the agent obtains all the keys and locks, falls into a pit, or
reaches 100 time steps.
A state in the key-lock environment is represented as a vector, including the Euclidean distance from the
learner in all cardinal directions to the nearest key and lock, four binary
parameters indicating if there is an obstacle in the neighboring cells, four binary parameters indicating if there is a key or lock in the neighboring cells, and eight binary parameters indicating if there is a pit in the two adjacent cells in all four directions. Lastly, two integers indicate the number of keys and locks captured so far. We use the dual DQN architecture to train on Key-Lock environments since the domain has a discrete state-action space.

\subsubsection{Flags}

The second domain
contains flags.  The agent's task is to capture the flags in an order unknown to it prior to training. Each flag picked up gives
a reward of 10 and the reward value increases by an additional 10 for every consecutive flag the agent captures.  All other actions including moving into an obstacle receive
-10. The learner can
only move in cardinal directions and is assumed to have obtained the flag if its location matches the location of the flag.  An episode
terminates when the agent obtains all the flags, or
reaches 100 time steps.
A state in the capture-the-flag environment is represented as a vector, including the Euclidean distance from the
learner in all cardinal directions to all of the flags, four binary
parameters indicating if there is an obstacle in the neighboring cells, and four binary parameters indicating if there is a flag in the neighboring cells. Lastly there is one integer for the number of flags captured so far. We use the dual DQN architecture to train on capture-the-flag environments since the domain has a discrete state-action space.

\subsubsection{Parking}

The last domain is the parking environment from \citet{highway-env}. The parking lot consists of 30 parking spots, an agent and a goal item randomly positioned in one of the spots at the beginning of each episode. The agent always starts in the same position but its initial orientation changes randomly. The agent's task is to reach the goal and orient itself in the right direction. The agent takes two continuous actions: velocity and angular velocity both defined in the range of [-1,1]. We use the actor-critic architecture to train on Highway-Parking environment because the domain has a continuous state-action space. Since the output of the model is continuous in this domain, we assume a normal distribution for the output and use the relative entropy for continuous Gaussian variables in calculating the uncertainty, which has a closed form solution for normal distributions given in \citet{Pardo2005StatisticalIB}. The environment allows the agent to wander outside of the parking lot, which increases the size of the state space. The agent receives a punishment proportional to its distance to the goal at every step of the training. An episode terminates when the agent attains the goal position with the proper heading or when it reaches 100 time steps.
A state in the parking environment is a vector, including the agent's position, velocity, angular velocity, and the goal's position and orientation. The input to the neural network is the concatenation of the state and goal information since the environment uses a goal-aware observation space. We use 

\subsubsection{Environments for Regressor Training}

For the training of the regressor, we use simplified versions of the target tasks. In the case of key-lock and capture-the-flag domains, the regressor learns a $10\times10$ environment having the same goal characteristics (position and direction) as the original environment. In the parking environment, we reduce the number of parking spots to 8 and halve the initial distance between the agent and the parking spaces. Since \readcsa{} requires the training of the regressor for a particular domain once, we only need to construct a single source environment for every domain, which ends up being less time-consuming than manual curricula generation.

For comparison, we also train different regression models on different source tasks to see how the regression training affects the curriculum performance. We use three settings: source task simpler than the target, source task similar to the target and source task same as the target. Source task simpler than the target denotes the $10\times10$ environment we describe in section \ref{test}, source task similar to target uses a $20\times20$ environment with slightly different goal positions than the target and source task same as target uses the target environment in training the regressor for the Key-Lock and Capture the Flag domain. Similarly, source task simpler than target uses half the number of parking spots with half the initial distance to the agent, source task similar to target uses similar number of parking spots to the target task and source task same as target uses the target environment in training the regressor for the Parking domain. 

\subsection{Comparison Algorithm}

We compare \readctd{} and \readcsa{} to the method
from~\citet{NarvekarPolicyChange}, which uses the maximum change in the policy as
determined by the number of states in which the action selected by the policy
before learning the source task differs from the action selected by the policy
after learning the source task. The criteria selects the task with the highest
policy change to the curriculum. We use a set of 15 manually generated for this
comparison algorithm.

For additional comparison, we also use a randomly generated curriculum and
learning directly on the target task. We generate the random curriculum by
sampling start states from the neighborhood of the terminal states.

\subsection{Experimental Setup and Hyperparameters}

We compare performance based on the total reward obtained while learning the
target and curriculum tasks. We also compare the convergence times and
provide 95\% confidence intervals. All algorithms share the same target task.

There are two types of hyperparameters: 1) neural network (including for the
convergence criteria) and 2) curriculum selection. Table~\ref{table:hyper}
contains the parameters used by the neural network in the key-lock environment. These parameters are not
tuned for any specific algorithm and all algorithms we use share these
parameters for all of the experiments and results in the same domain. $\epsilon$ is the
$\epsilon$-greediness of the algorithm, which decays during training. $\alpha$
and $\gamma$ are the learning rate and discount factor respectively. Entropy
Reduction is the number of consecutive episodes the convergence criteria
considers in determining if entropy is reducing or not.

Table~\ref{table:hyper3} shows the parameters for \readctd{}.  Clustering and
Cutoff Distance refer to the clustering algorithm used to generate regions and
the cutoff point for stopping the merge of clusters.  $\eta$ denotes the number
of training episodes on the target task before the curriculum generation.
Lastly, Curriculum Length is the number of tasks in the curriculum. For
\readcsa{}, $\beta$ is 15,000 and $\eta$ is $40,000$.

\begin{table}[tbh]
\centering

 \begin{tabular}{ c c | c c } 
 \toprule
\textbf{Param} & \textbf{Value} & \textbf{Param} & \textbf{Value} \\ [0.5ex] 
  \midrule
 $\epsilon$ & 1 &  Optimizer & Adamax \\

  $\epsilon$ Decay & 0.995 & Loss & Mean Squared \\

 Minimum $\epsilon$ & 0.01 & Buffer Size & 40,000 \\ 
 
 $\alpha$ & 0.005 & Batch Size & 16 \\

 $\gamma$ & 0.99 & Entropy Reduction & 10 \\ 

\bottomrule
\end{tabular}
\caption{Hyperparameters for the neural network.}
\label{table:hyper}
\end{table}

\begin{table}[tbh]
\centering
 \begin{tabular}{ c c | c c } 
 \toprule
\textbf{Parameter} & \textbf{Value} & \textbf{Parameter} & \textbf{Value} \\ [0.5ex] 
  \midrule
   Clustering & Agglomerative  &  Cutoff Distance & 3 \\ 
   $\eta$ & 50,000 &  Curriculum Length & 4 \\ 
\bottomrule
\end{tabular}
\caption{Hyperparameters for the curriculum learning.}
\label{table:hyper3}
\end{table}

\section{Results and Analysis}\label{result}

We perform 25 runs for each algorithm and report the average.  All algorithms
use the same randomly initialized weights and we alter the DQN weights every run.

\subsection{Results for Key-Lock Domain}

Here we present the evaluation results and analysis for the Key-Lock Domain. Without specification, the default training setting for \readctd{} is without any heuristic distance measures. The default training setting for \readcsa{} is 200 clusters and the simpler-than-target source task. READ-C-SA with No Clusters directly uses the visited states in curriculum generation. Other than the comparison of performance under different target environment complexity, the default target environment is the $20\times20$ environment described in section~\ref{test}.

\begin{figure*}[htb]
    \centering
    \subfigure[Curriculum + Target Performance for \readc{} and Baselines]{\includegraphics[width=0.49\columnwidth]{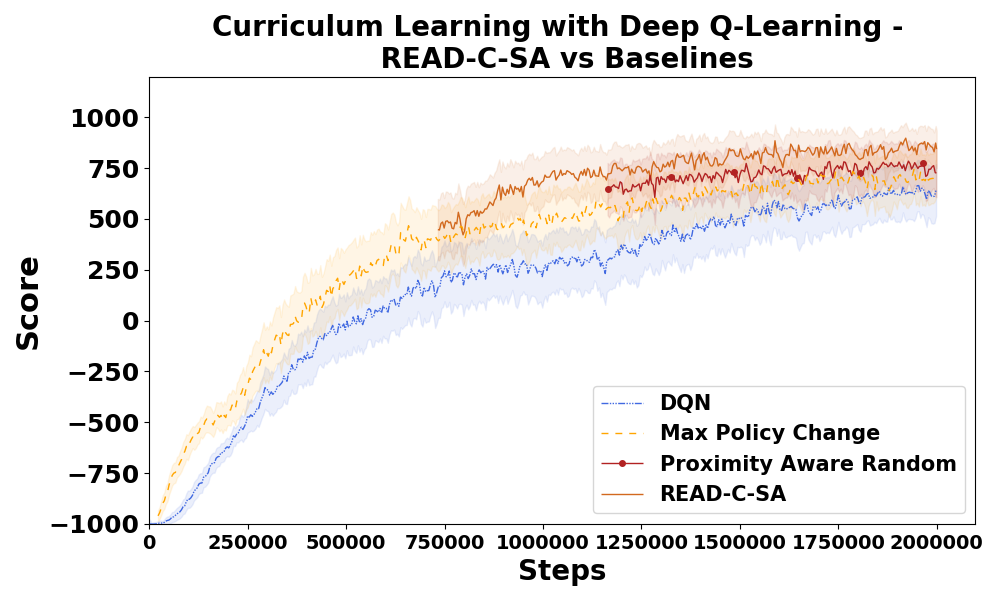}\label{fig:fig1}}
    \subfigure[Curriculum + Target Performance + Generation Overhead for \readc{} and Baselines]{\includegraphics[width=0.49\columnwidth]{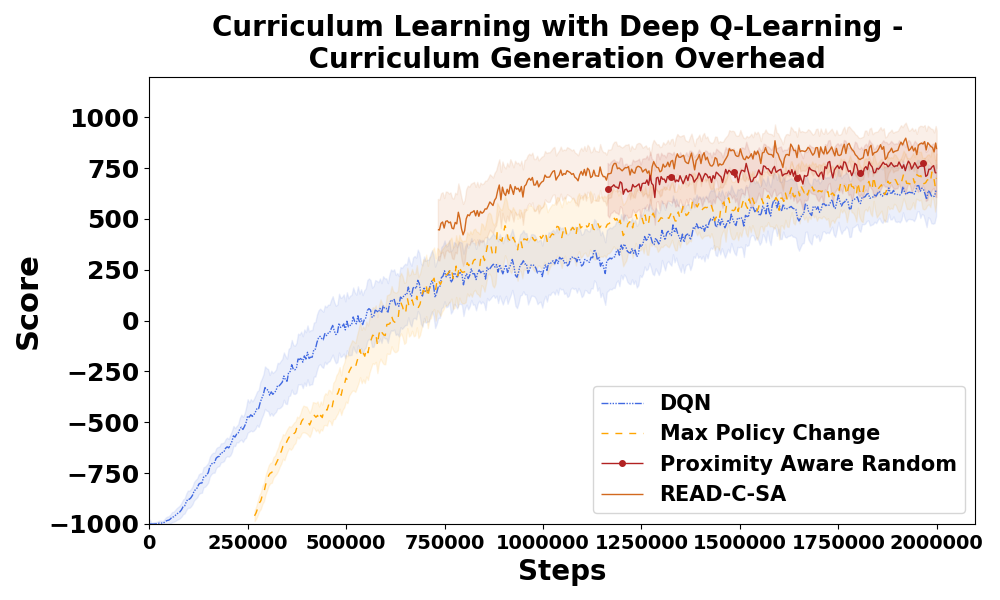}\label{fig:fig2}}
    \subfigure[Curriculum + Target Performance for Variants of \readc{}]{\includegraphics[width=0.49\columnwidth]{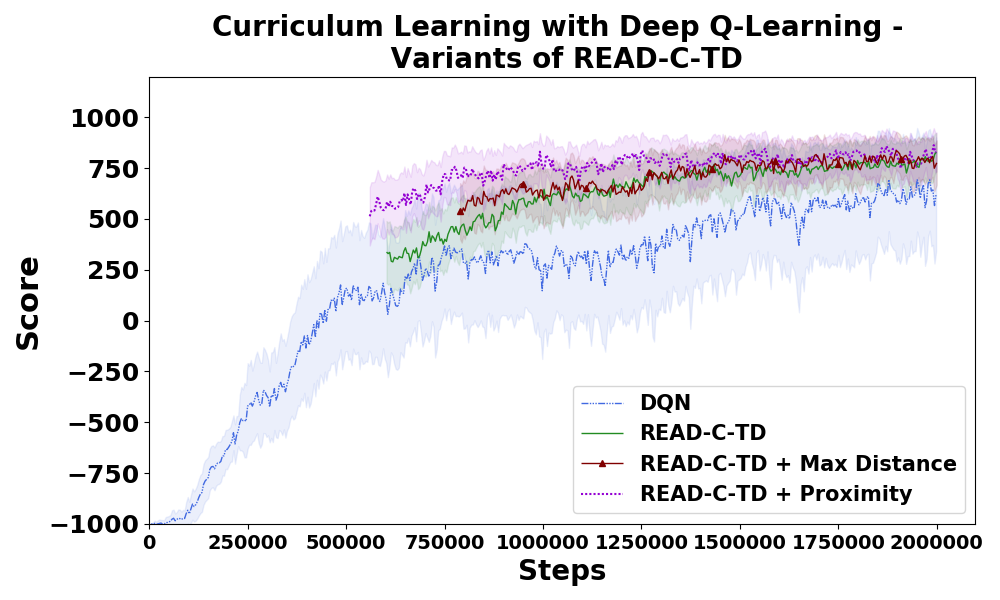}\label{fig:fig3}}
    \subfigure[Curriculum + Target Performance for \readctd{} and \readcsa{}]{\includegraphics[width=0.49\columnwidth]{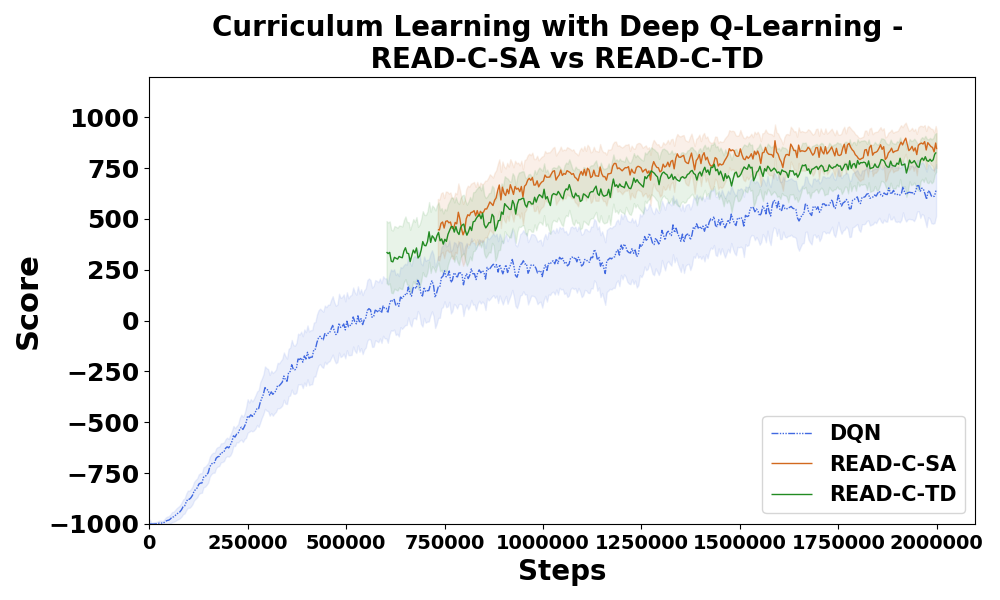}\label{fig:fig4}}
    \subfigure[Curriculum + Target Performance for Different Regressors]{\includegraphics[width=0.49\columnwidth]{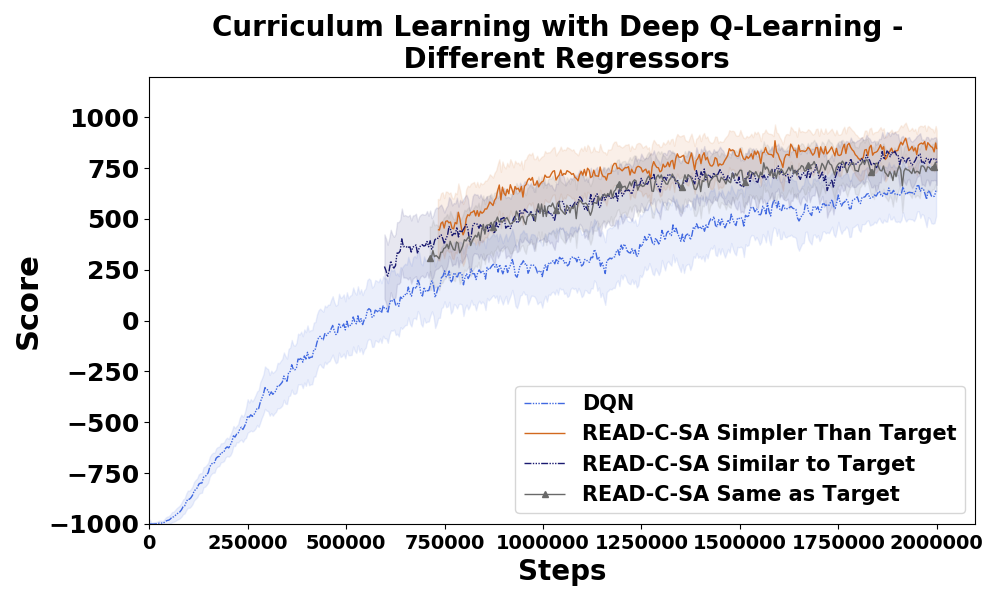}\label{fig:fig5}}
    \caption{Performance of the Curriculum-Learning Algorithms as a Function of Training Steps in Key-Lock Domain.}
    \label{fig:fig6}
\end{figure*}

\begin{figure*}[htb]
    \centering
    \subfigure[Cluster Size Comparison for 10 $\times$ 10 Target Environment]{\includegraphics[width=0.49\columnwidth]{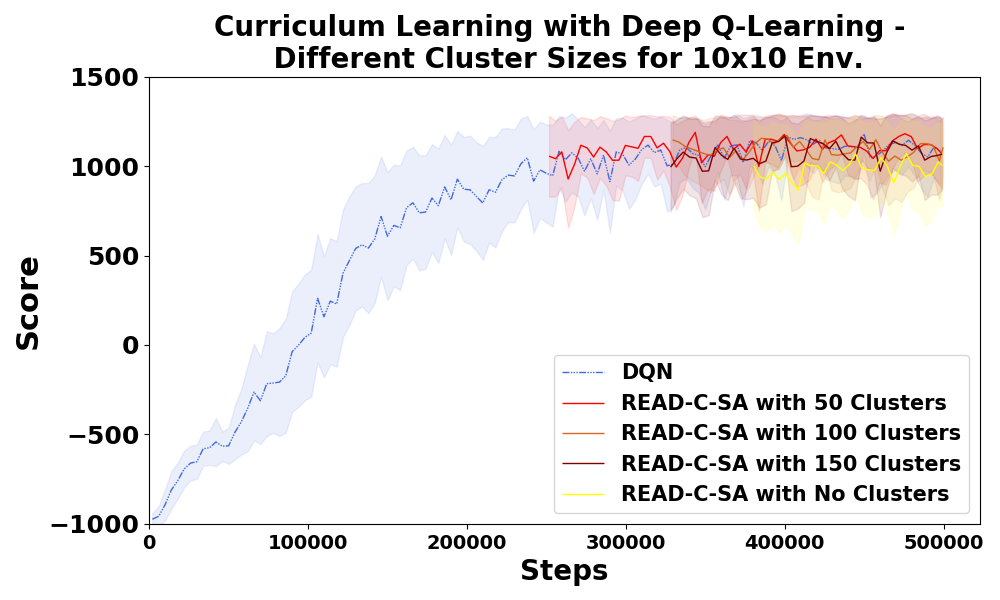}\label{fig:fig7}}
    \subfigure[Cluster Size Comparison for 20 $\times$ 20 Target Environment]{\includegraphics[width=0.49\columnwidth]{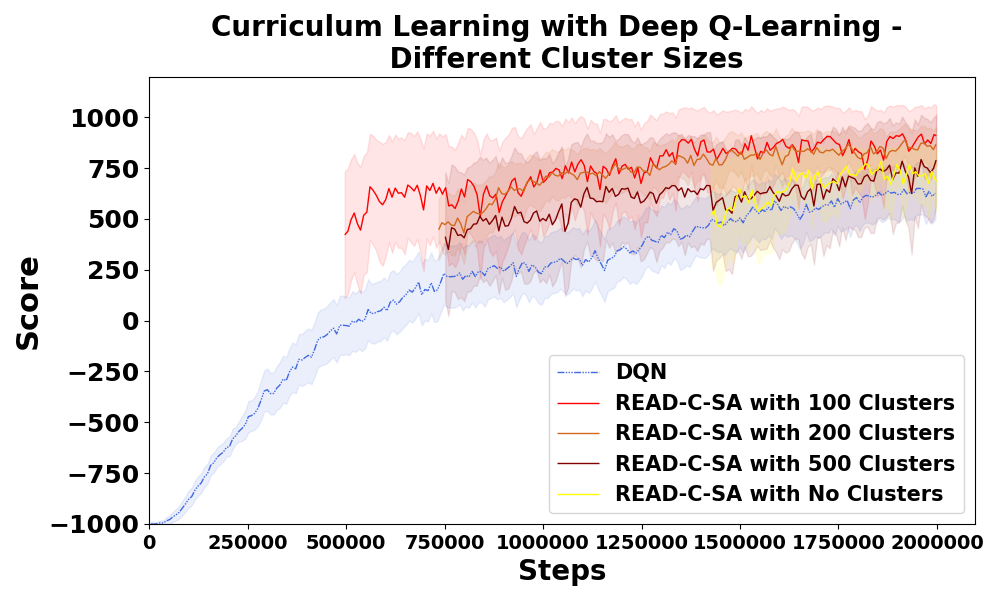}\label{fig:fig8}}
    \subfigure[Cluster Size Comparison for 30 $\times$ 30 Target Environment]{\includegraphics[width=0.49\columnwidth]{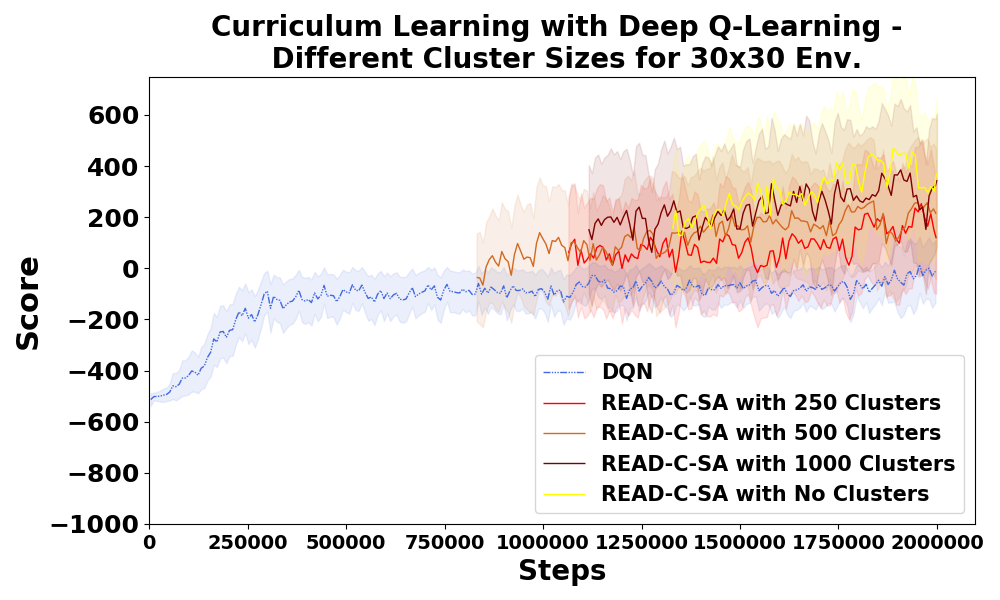}\label{fig:fig9}}
    \caption{Effect of Cluster Size on the Performance of the Curriculum-Learning Algorithms in Key-Lock Domain.}
    \label{fig:fig10}
\end{figure*}

Figure~\ref{fig:fig1} contains results for \readcsa{} in relation to the Max
Policy Change and DQN baselines in the key-lock domain as a function
of the number of training steps. Figure~\ref{fig:fig2} displays the same
results while also counting the curriculum-generation overhead. Only Max Policy Change algorithm has curriculum generation overhead. Following the procedure from the original paper of Max Policy
Change \citep{NarvekarPolicyChange}, we perform 50,000 iterations of training on the target task to generate
the prior policy and then perform 5,000 iterations of training on each source
task to obtain the posterior policy used to measure the policy change. We
repeat this process for two curriculum steps which, in total, results in
$2 \times (50,000 + 15 \times 5,000) - 5,000 = 245,000$ iterations of curriculum-generation
overhead. \readc{} and the other baselines, on the other hand, does not require us to perform any
training for the curriculum generation. Each graph is offset by the amount of time required to do the curriculum training.
Figure~\ref{fig:fig3} compares \readctd{}'s different
heuristic variants whereas Figure~\ref{fig:fig4} shows the performance of \readcsa{} with respect to \readctd{} and no curriculum baseline. Figure~\ref{fig:fig5} displays the results obtained from training the regressor on data coming from three different source tasks. Figures~\ref{fig:fig10} compares the performance of \readcsa{} for different cluster numbers on varying sizes of target environment.  Shaded regions are 95\% confidence intervals. 

\readcsa{} converges faster than the baseline algorithms even if we do not
account for the curriculum-generation overhead but its confidence intervals
also show an overlap with max policy change (Figure~\ref{fig:fig1}). Accounting
for curriculum-generation overhead, the overlap between \readctd{} and Max
Policy Change reduces notably indicating performance improvements from
\readcsa{} (Figure~\ref{fig:fig2}). \readcsa{} and proximity aware random curricula perform
similar to one another but \readcsa{} results in much shorter curriculum
training (Figure~\ref{fig:fig2}). Furthermore, \readctd{} + proximity performs better than \readctd{}, and \readctd{} + max distance, showing that heuristic
functions in combination with relative entropy can produce higher performance
(Figure~\ref{fig:fig3}). \readctd{} performs slightly worse than
\readcsa{} (Figure~\ref{fig:fig4}) mainly due to the fact that a regressor trained on simpler environments generalizes better and captures more relevant parts of the agent's uncertainty compared to a teacher that calculates uncertainty for a larger state space. Training the regressor on a task Similar to Target or Same as the Target results in a performance close to \readctd{} slightly below the performance of \readcsa{} with Simpler Than Target (Figure~\ref{fig:fig5}), which is not unusual since \readctd{} uses the training data on the target task to generate the curriculum as well. 

All \readc{} approaches in a small environment give similar results despite the difference in cluster sizes due to the target environment being simple to learn  (Figure~\ref{fig:fig7}). Moderate number of clusters such as 100 or 200 performs the best for medium size environments (Figure~\ref{fig:fig8}) while no clustering gives the best results followed closely by 1000 clusters in the large environment although the performance for 500 clusters is very close to 1000 clusters as well (Figure~\ref{fig:fig9}). In general, it is safe to say that using a moderate number of clusters always guarantees good performance for \readcsa{} on any size of environment even though it might not always produce the best results. Here please note that we use the same set of distance cut-off thresholds ([0.1, 1, 3, 5]) for Agglomerative Clustering in all of these environments but depending on the size of the environment, the number of clusters being generated differs from one figure to the other, which is why the number of clusters is higher in the case of large environment compared to the moderate-size environment (Figures~\ref{fig:fig8}~\&~\ref{fig:fig9}).

\begin{figure*}[htb]
    \centering
    \subfigure[The Step Number Where Each Algorithm Reaches A Reward of 900 for 10 Consecutive Episodes]{\includegraphics[width=0.8\columnwidth]{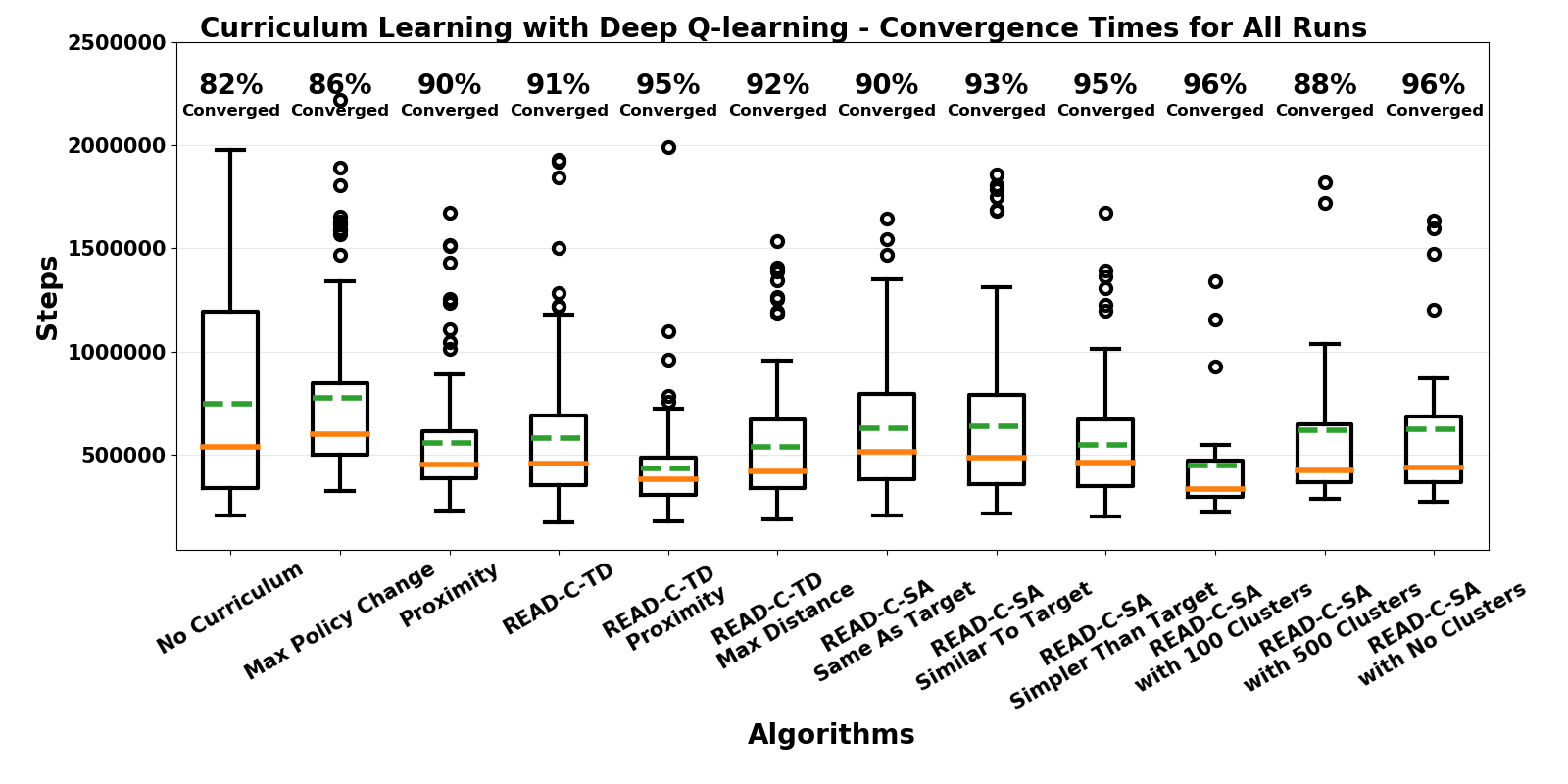}\label{fig:fig11}}
    \subfigure[The Step Number Where Best Performing 80\% of the Runs Reach A Reward of 900 for 10 Consecutive Episodes]{\includegraphics[width=0.8\columnwidth]{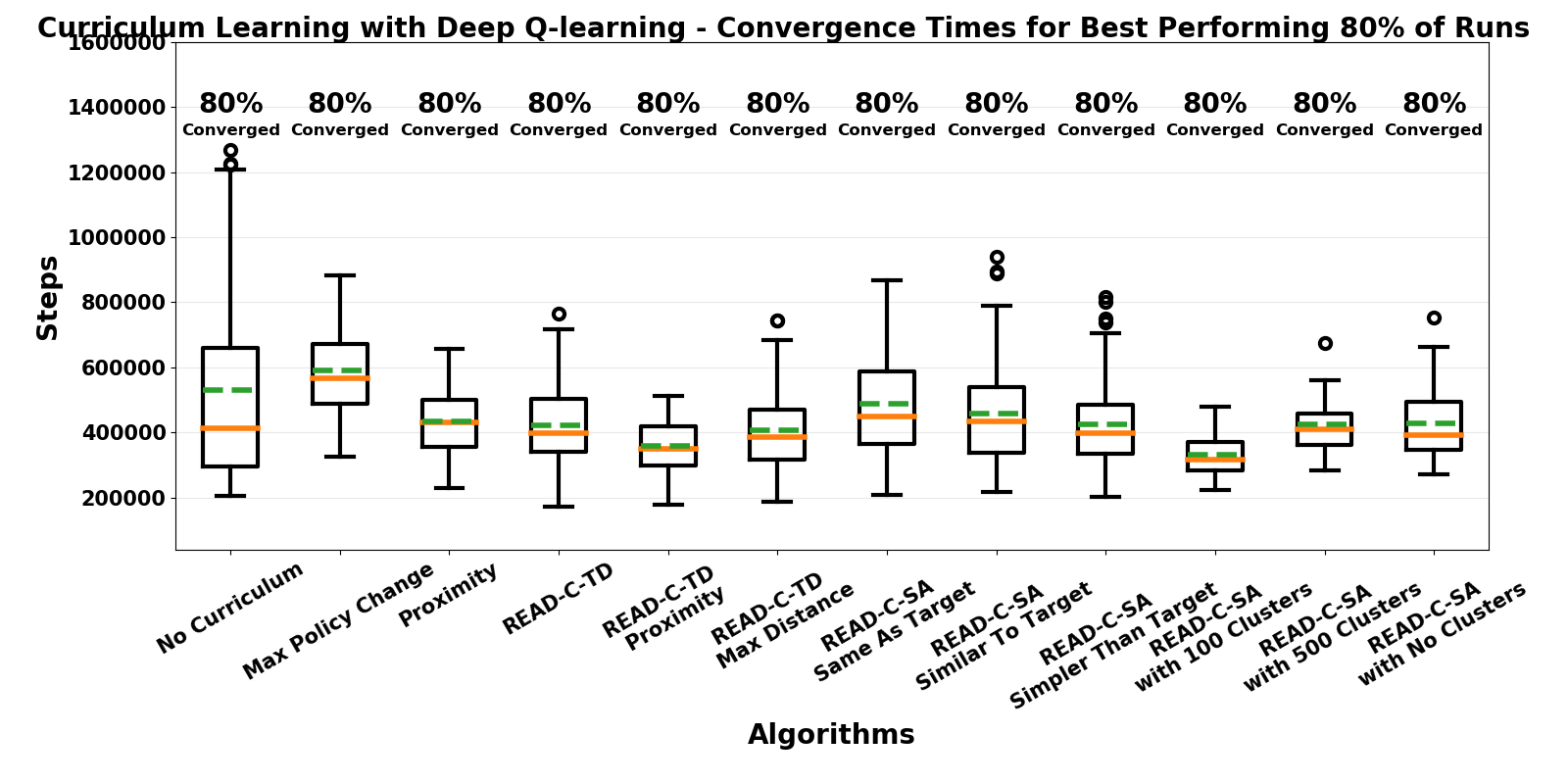}\label{fig:fig12}}
    \caption{ Box Plots for the Convergence Times of the Algorithms in Key-Lock Domain.}
    \label{fig:fig13}
\end{figure*}

Convergence results are presented as box plots illustrating the distribution of
the number of steps to reach a given cumulative reward across 25 runs
(Figure~\ref{fig:fig13}). The orange solid line is the median and green dotted
line the mean. Each graph represents the total convergence time for curriculum
and target training. Runs that do not reach highest reward at any point in the
execution are not included in the box plots since there is no convergence time to include for those runs. However, above each box, we also include the
percentage of runs that converge to the highest cumulative reward, which tells us how likely it is for the runs of an algorithm to reach the highest cumulative reward during execution. If the convergence rate above the graph is low, it means that the given algorithm has a lot of runs that fail to converge. 

Figure~\ref{fig:fig12} results are selected from the fastest converging 80\% of
runs so the graph contains the same number of runs for every algorithm.

\readcsa{} with 100 clusters has the lowest variance in the convergence times and
highest rate of convergence (Figures~\ref{fig:fig11}~\&~\ref{fig:fig12}). \readctd{} and its variants generally result in better mean and median convergence times than the
baseline algorithms whereas \readcsa{} approaches generate varying performance
results depending on how many clusters are used for the curriculum generation. \readcsa{} with 100 clusters using the simpler source environment produces the lowest mean
convergence time followed closely by \readctd{} + proximity. \readcsa{} with 100 clusters also shows only a slightly higher
convergence rate than \readctd{} and its variants arguably making it the best converging variant of \readc{} (Figure~\ref{fig:fig11}). On the other hand, \readcsa{} with source tasks similar to target and same as target show worse mean and median performance than \readctd{} and proximity aware baseline. The problem in these approaches stem from the regressor training on larger state spaces with higher variance and error rates, which consequently harms the performance on the target environment and shows the importance of using a suitable small environment in modeling the uncertainty of the agent in a particular domain. If we only look at 80\% of
the runs, the difference between median and mean
performance clearly reduces for many of the best performing \readc{}
algorithms, indicating that the outliers are influencing the
mean convergence times being higher than medians (Figure~\ref{fig:fig12}). \readcsa{} with 100 clusters still performs the best when looking at 80\% of
the runs and the relative ordering of the algorithms mostly remains the same as before.

\subsection{Results for Capture-the-Flag Domain}
Here we provide the evaluation results and analysis for Capture-the-Flag domain. The default cluster size for \readcsa{} is 150. Other default settings are the same as results for Key-Lock domain.

\begin{figure*}[htb]
    \centering
    \subfigure[Curriculum + Target Performance for \readc{} and Baselines]{\includegraphics[width=0.49\columnwidth]{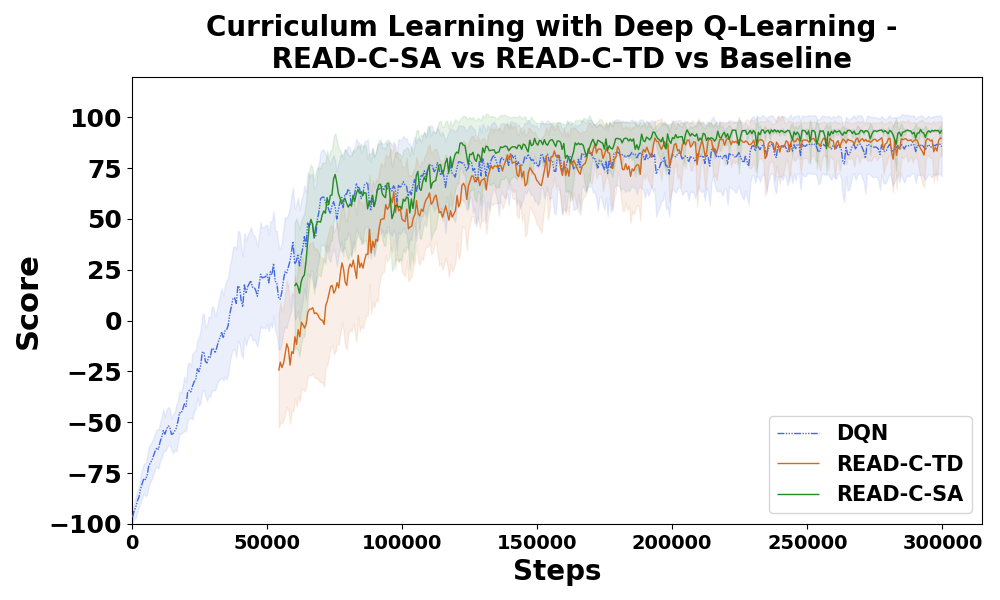}\label{fig:fig14}}
    \subfigure[Curriculum + Target Performance for Variants of \readc]{\includegraphics[width=0.49\columnwidth]{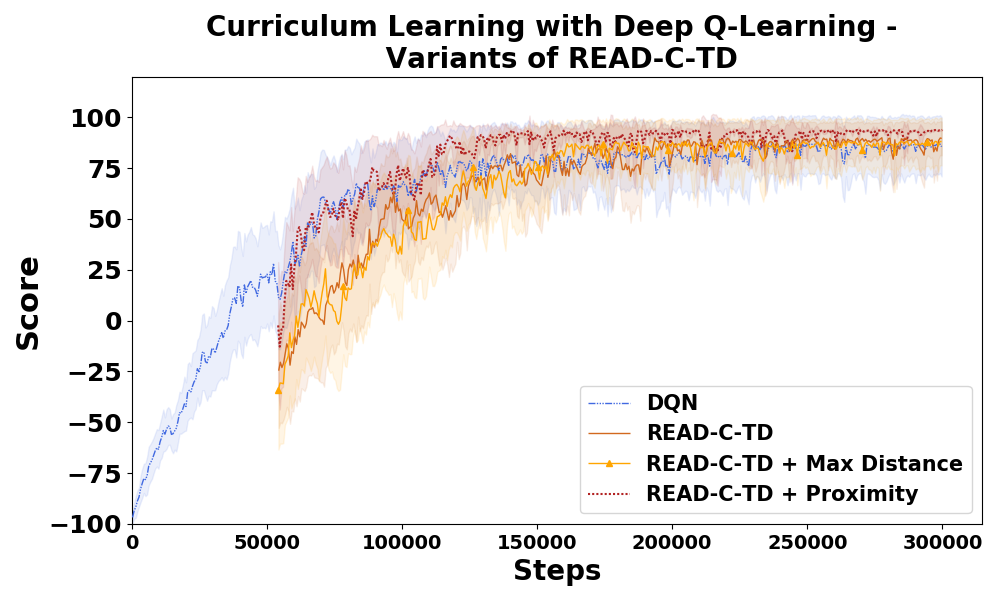}\label{fig:fig15}}
    \subfigure[Curriculum + Target Performance for Different Regressors]{\includegraphics[width=0.49\columnwidth]{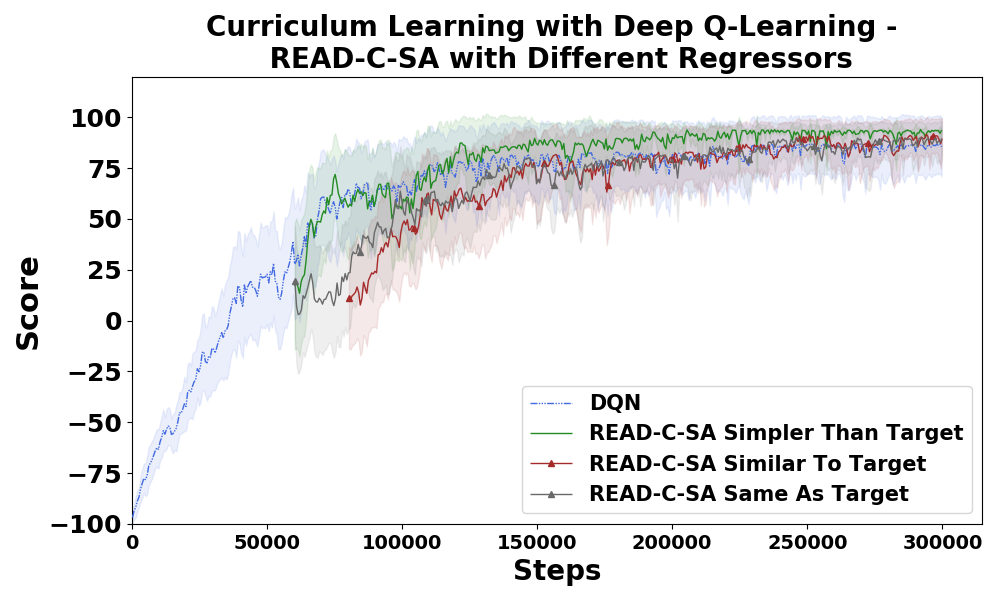}\label{fig:fig16}}
    \caption{Performance of the Curriculum-Learning Algorithms as a Function of Training Steps in Capture-the-Flag Domain.}
    \label{fig:fig17}
\end{figure*}

\begin{figure*}[htb]
    \centering
    \subfigure[Cluster Size Comparison for 10 $\times$ 10 Target Environment]{\includegraphics[width=0.49\columnwidth]{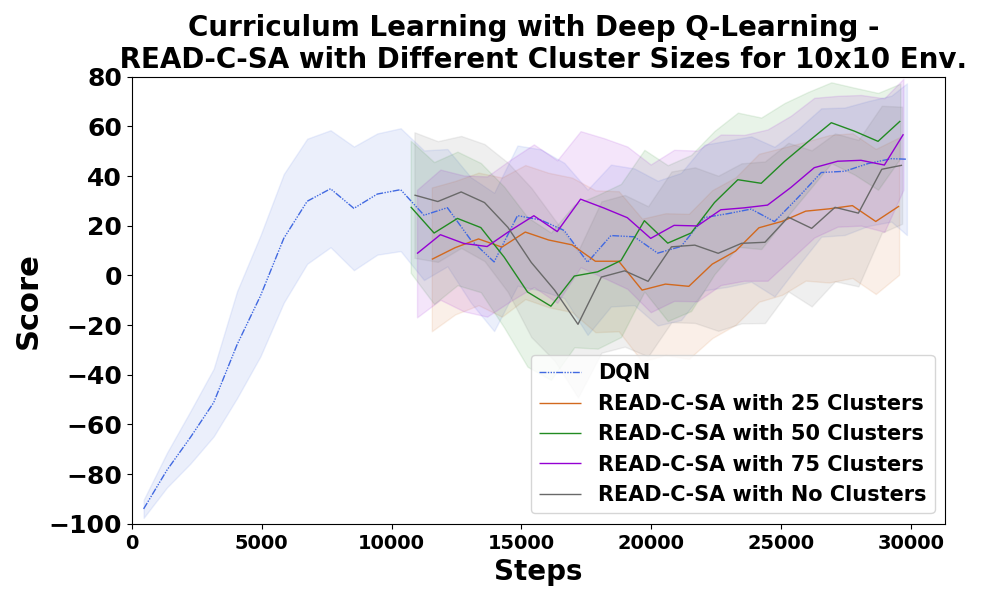}\label{fig:fig18}}
    \subfigure[Cluster Size Comparison for 20 $\times$ 20 Target Environment]{\includegraphics[width=0.49\columnwidth]{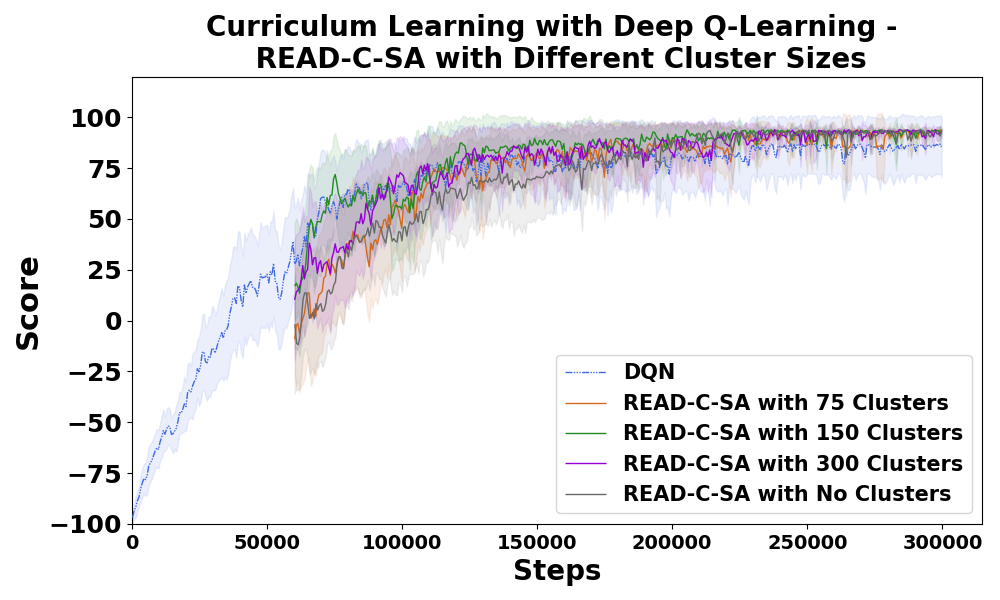}\label{fig:fig19}}
    \subfigure[Cluster Size Comparison for 30 $\times$ 30 Target Environment]{\includegraphics[width=0.49\columnwidth]{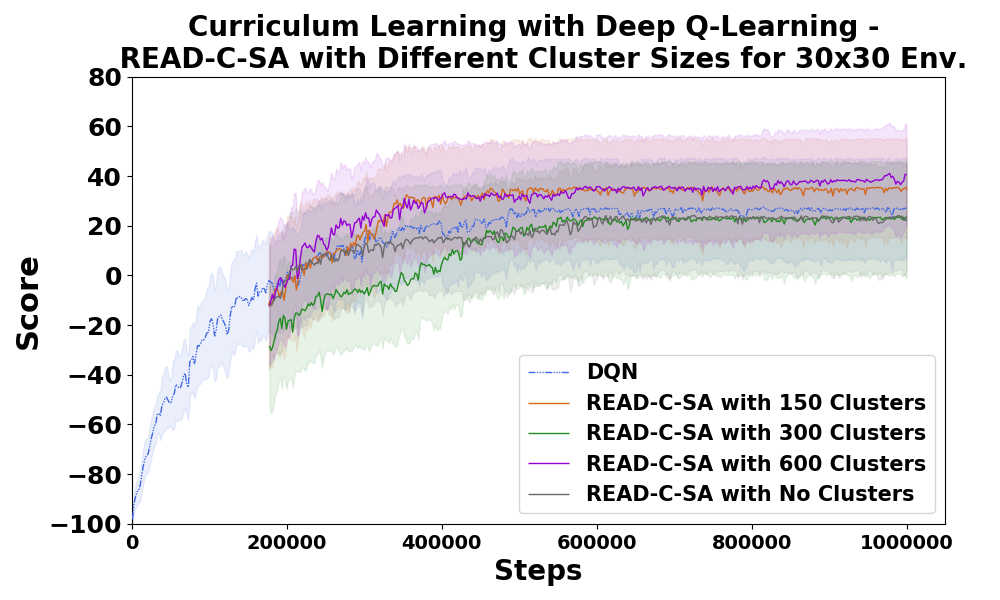}\label{fig:fig20}}
    \caption{Effect of Cluster Size on the Performance of the Curriculum-Learning Algorithms in Capture-the-Flag Domain.}
    \label{fig:fig21}
\end{figure*}

\begin{figure*}[htb]
    \centering
    \subfigure[The Step Number Where Each Algorithm Reaches A Reward of 84 for 500 Consecutive Episodes]{\includegraphics[width=0.95\columnwidth]{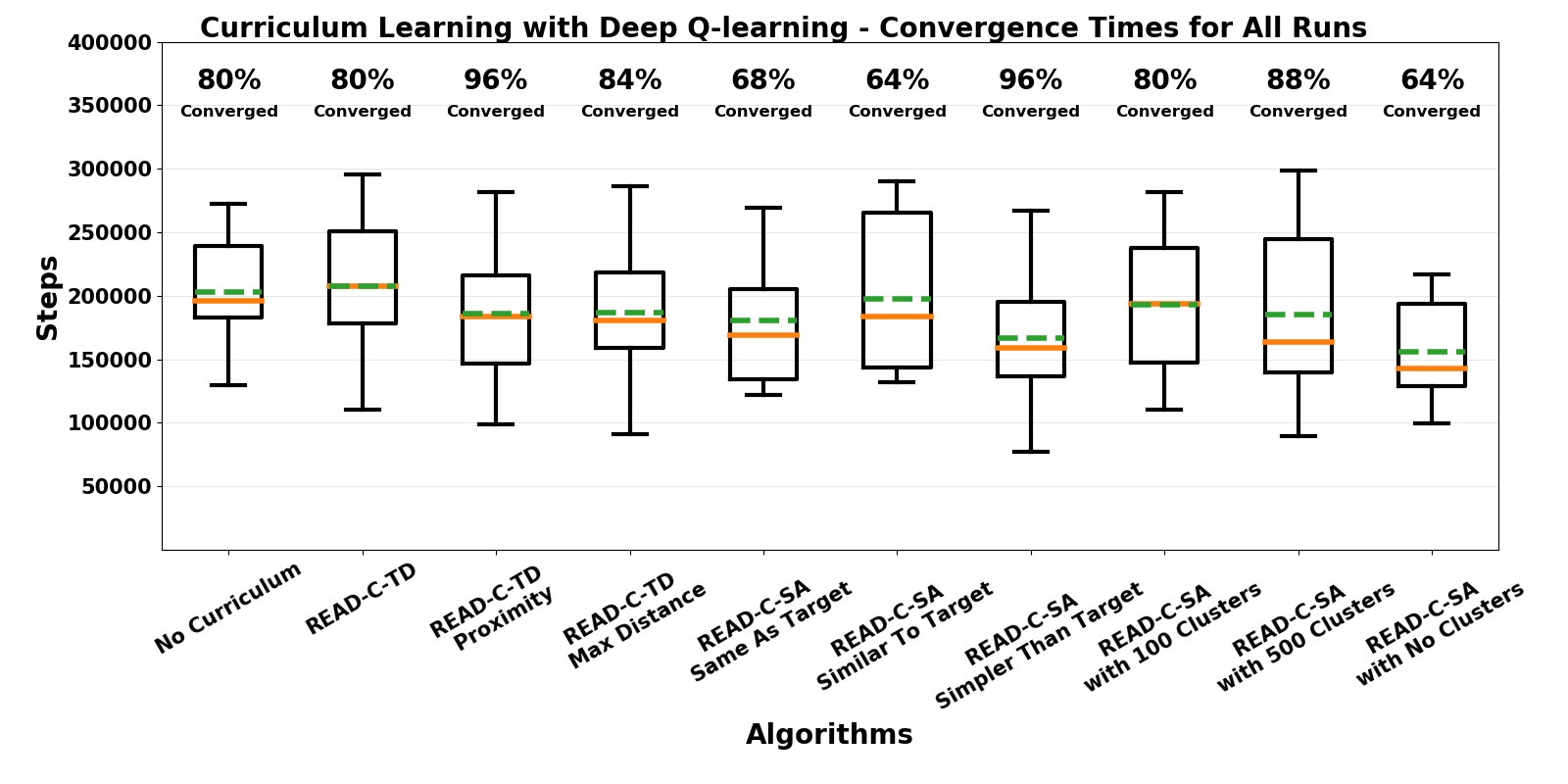}\label{fig:fig22}}
    \subfigure[The Step Number Where Best Performing 80\% of the Runs Reach A Reward of 75 for 500 Consecutive Episodes]{\includegraphics[width=0.95\columnwidth]{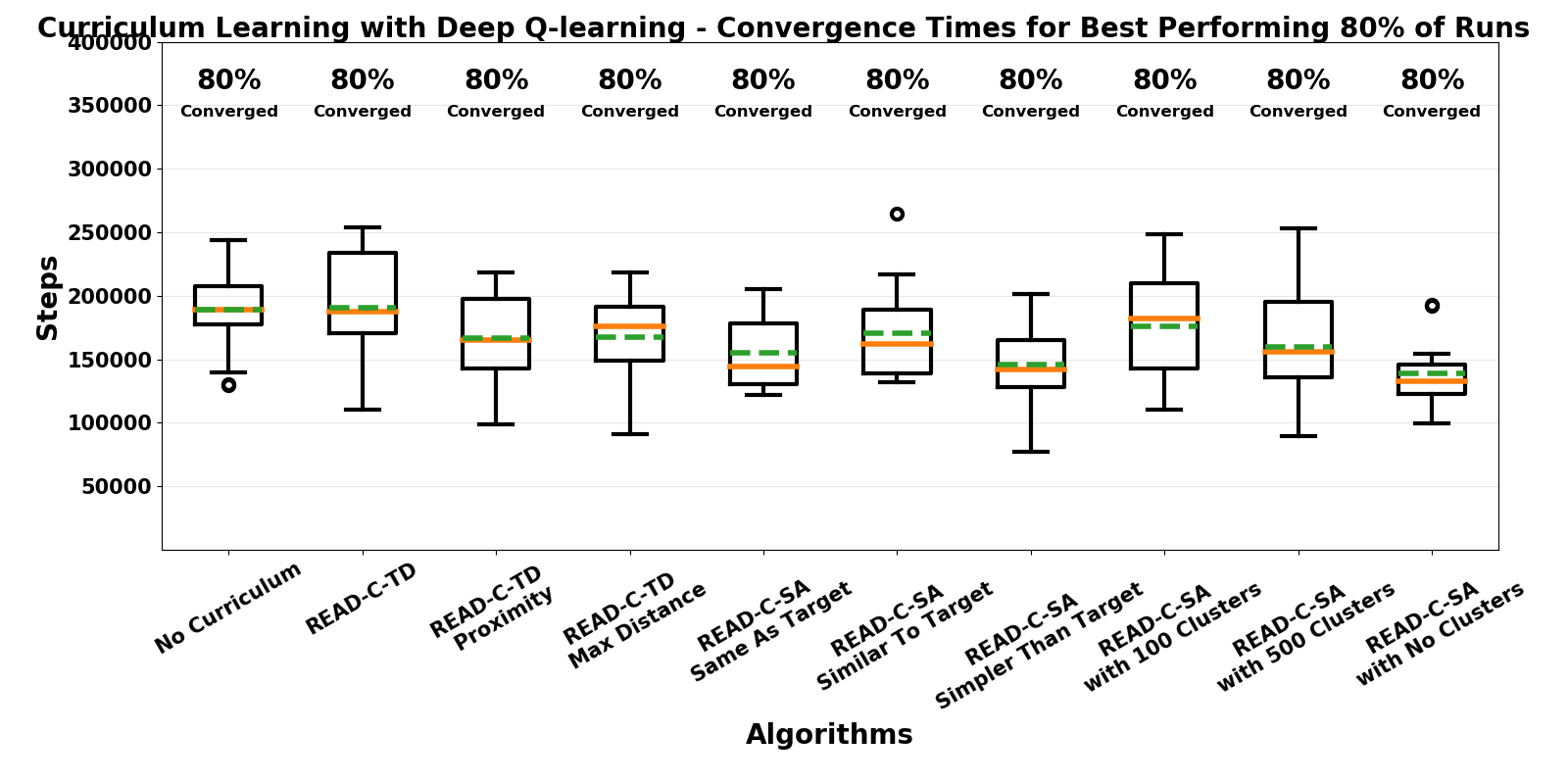}\label{fig:fig23}}
    \caption{ Box Plots for the Convergence Times of the Algorithms in Capture-the-Flag Domain.}
    \label{fig:fig24}
\end{figure*}

Figure~\ref{fig:fig14} shows the comparison of \readcsa{}, \readctd{} and no curriculum baseline. Figure~\ref{fig:fig15} shows the variants of \readctd{} and Figure~\ref{fig:fig16} displays different regressors trained on a $10\times10$ environment, a $20\times20$ environment similar to target task and the target environment respectively, on the capture-the-flag domain. Figure~\ref{fig:fig18}~,~Figure~\ref{fig:fig19}~\&~Figure~\ref{fig:fig20} display the effect of different cluster numbers on the performance of \readcsa{} using different target environment sizes. Finally, we present the box plots for convergence times of \readc{} approaches on the capture-the-flag domain illustrating the distribution of
the number of steps required to reach a given cumulative reward for a given number of consecutive episodes similar to the key-lock domain (Figure~\ref{fig:fig22}~\&~Figure~\ref{fig:fig23}).  Shaded regions show 95\% confidence intervals. 

\readcsa{} converges faster than the no curriculum baseline and \readctd{} (Figure~\ref{fig:fig14}). \readctd{} + proximity performs better than other \readctd{} variants (Figure~\ref{fig:fig15}). Using a simpler source environment for the regressor results in better performance compared to other types of regressors (Figure~\ref{fig:fig16}). The results are mainly consistent with what we have seen on the key-lock domain and further supports the evidence seen in the prior graphs. When it comes to the different cluster numbers, there does not emerge a meaningful difference in the performance of the algorithms in the small environment due to the environment taking little time to train despite the moderate cluster sizes reaching a slightly higher convergence point than the other options (Figure~\ref{fig:fig18}). For the moderate-size environment, using a moderate number of clusters such as 150 seems to result in the best performance although most of the cluster sizes show similar results in this domain (Figure~\ref{fig:fig19}). For the large-size environment, a moderate number of clusters between 159 and 600 again produces good performance similar to the key-lock domain (Figure~\ref{fig:fig20}). The choice of cluster numbers does not appear to have a large influence on the performance for capture-the-flag domain regardless of the target environment size and using any number of clusters in this experiment results in more or less slightly better outcome than the baseline approaches (Figures~\ref{fig:fig19}\&~\ref{fig:fig20}).

\readcsa{} with no clusters shows the lowest variance in the convergence times and produces the best mean and median performance while having a very low convergence rate meaning that many of its runs are classified as outliers (Figure~\ref{fig:fig24}). \readcsa{} with source task simpler than the target task gives the second best mean and median performance while also having the highest convergence rate (Figure~\ref{fig:fig22}), arguably making it the best performing algorithm in this domain. \readcsa{} with source task similar to the target task and same as the target task result in poorest convergence rates (Figure~\ref{fig:fig22}) and along with the results from the key-lock domain, prove that to obtain good performance on \readcsa{}, it is necessary to train the regressor on a simpler environment showing similarities to the target task. \readcsa{} with 500 clusters shows good median and mean performance similar to \readcsa{} with 200 clusters but it displays a higher variance among its runs indicating that using a too high number of clusters could make the algorithm more sensitive to the random weight initialization (Figure~\ref{fig:fig22}). If we only look at 80\% of
the runs, the variance among multiple executions clearly reduces for many of the best performing \readc{}
algorithms (Figure~\ref{fig:fig23}). Overall, the results for capture-the-flag domain show less variance (when accounted for the scale of the y-axis) and less performance differences among different \readc{} approaches compared to the key-lock domain, partially being caused by less-complex structure of the environment characteristics (Figure~\ref{fig:fig22}~\&~Figure~\ref{fig:fig23}).

\subsection{Results for Parking Domain}
Here we provide the evaluation results and analysis for the Parking domain. The cluster size remains the same for all \readcsa{} configurations. Other default settings are the same as the results from the grid environments.

\begin{figure*}[htb]
    \centering
    \subfigure[Curriculum + Target Performance for \readc{} and Baselines]{\includegraphics[width=0.49\columnwidth]{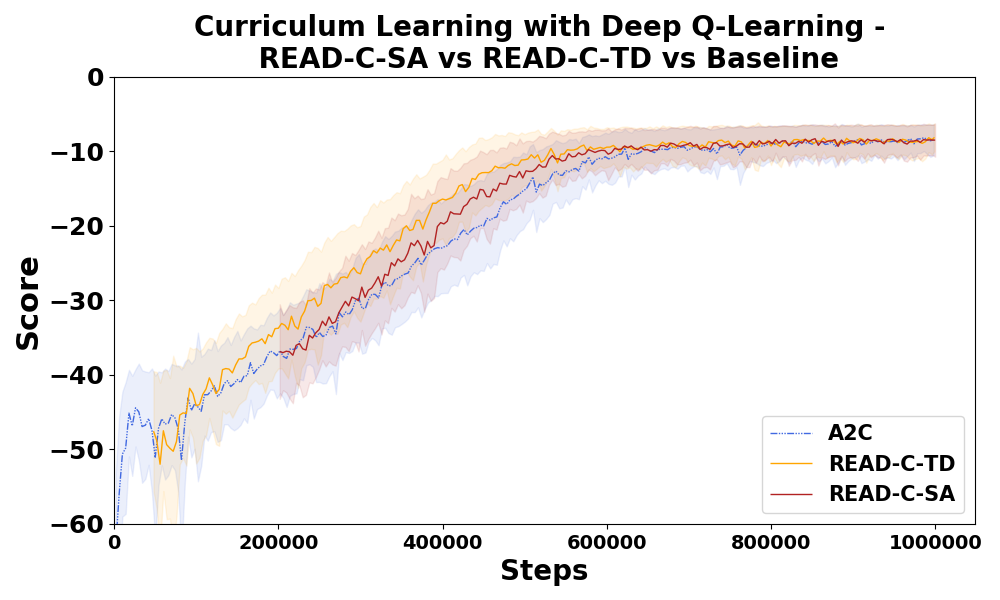}\label{fig:fig25}}
    \subfigure[Curriculum + Target Performance for Variants of \readc{}]{\includegraphics[width=0.49\columnwidth]{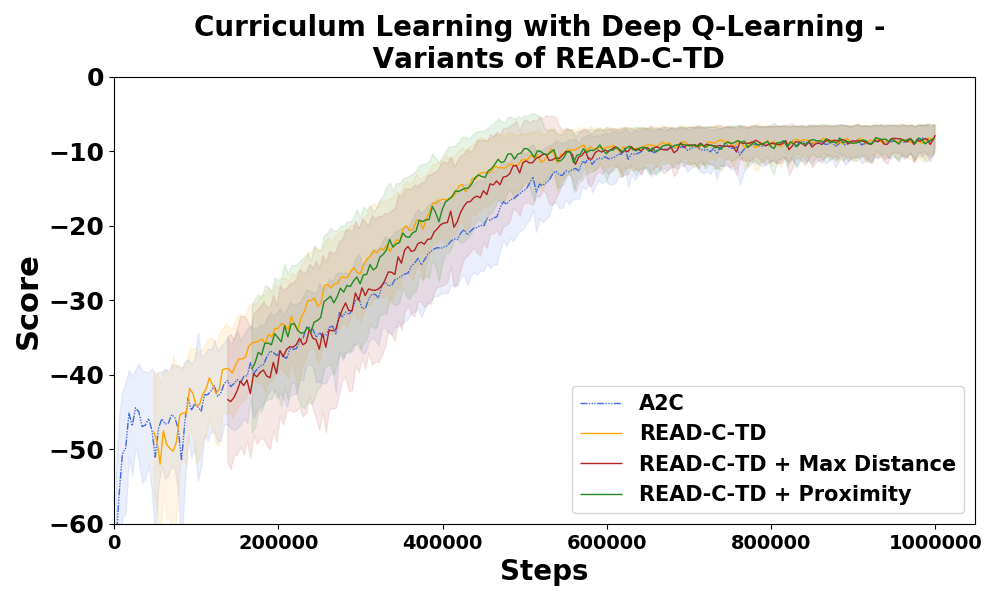}\label{fig:fig26}}
    \subfigure[Curriculum + Target Performance for Different Regressors]{\includegraphics[width=0.49\columnwidth]{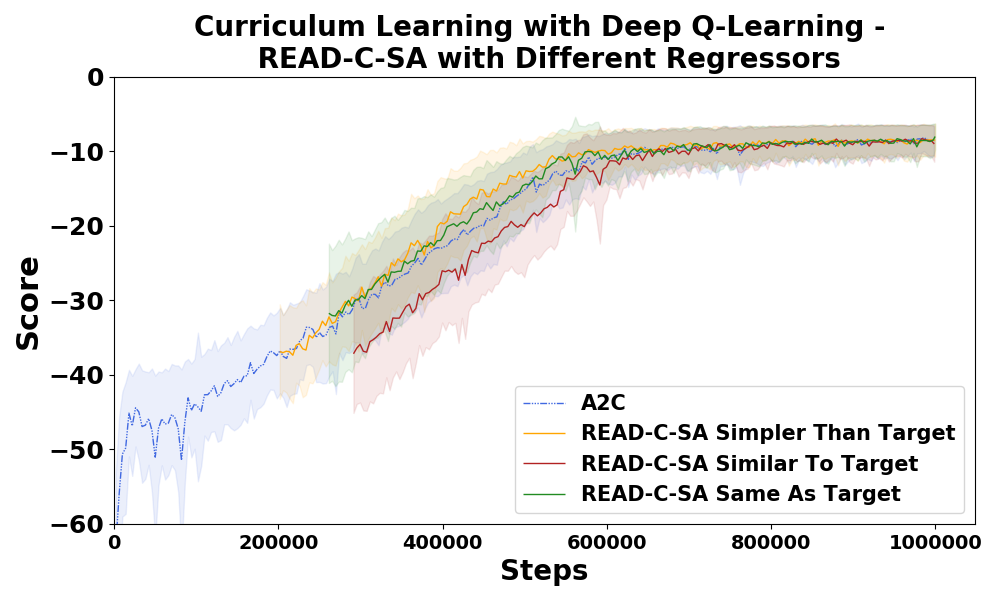}\label{fig:fig27}}
    \caption{Performance of the Curriculum-Learning Algorithms as a Function of Training Steps in the Parking Domain.}
    \label{fig:fig28}
\end{figure*}

\begin{figure*}[htb]
    \centering
    \subfigure[The Step Number Where Each Algorithm Reaches A Reward of -15 for 10 Consecutive Episodes]{\includegraphics[width=0.95\columnwidth]{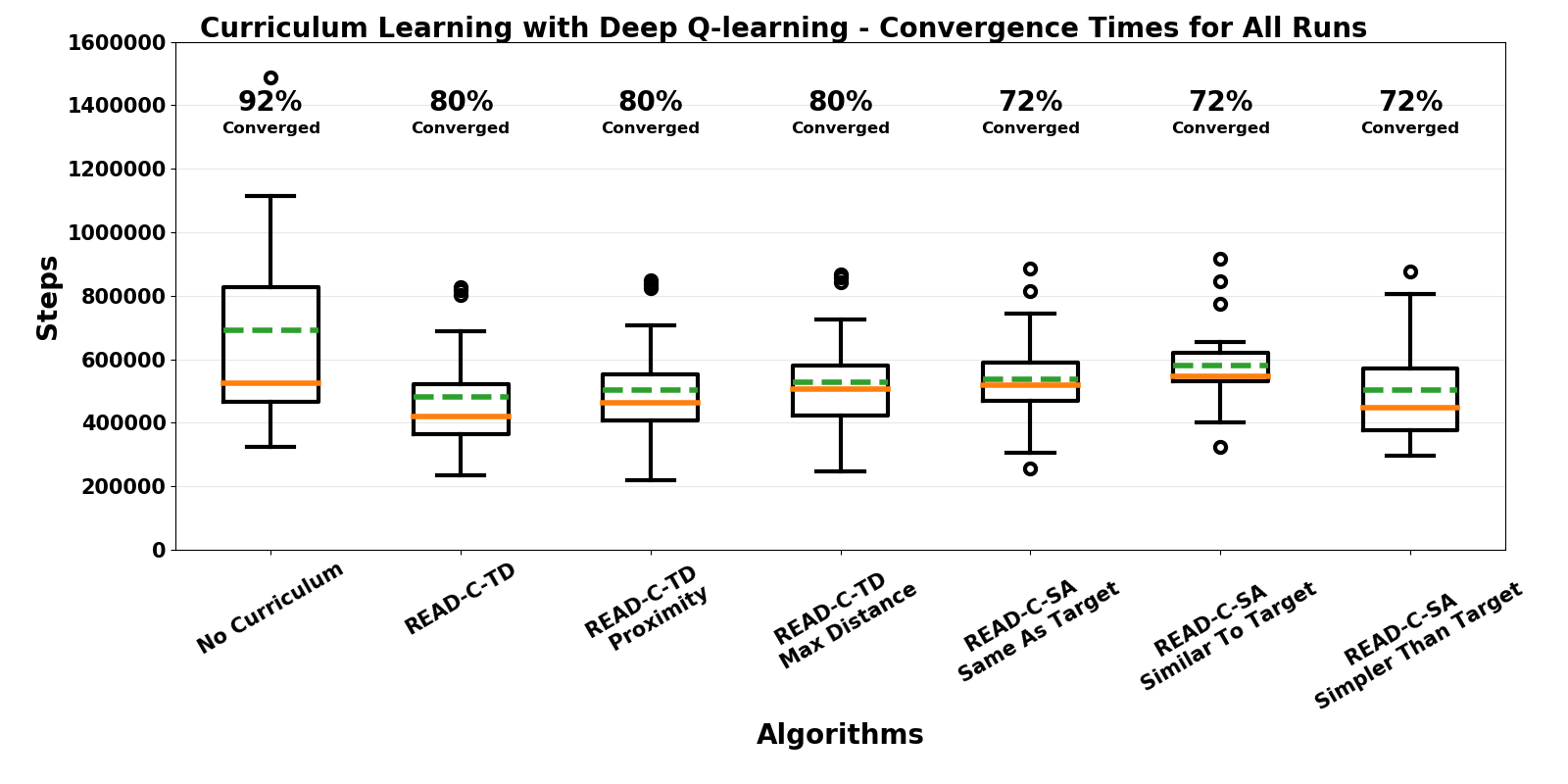}\label{fig:fig29}}
    \subfigure[The Step Number Where Best Performing 80\% of the Runs Reach A Reward of -20 for 10 Consecutive Episodes]{\includegraphics[width=0.95\columnwidth]{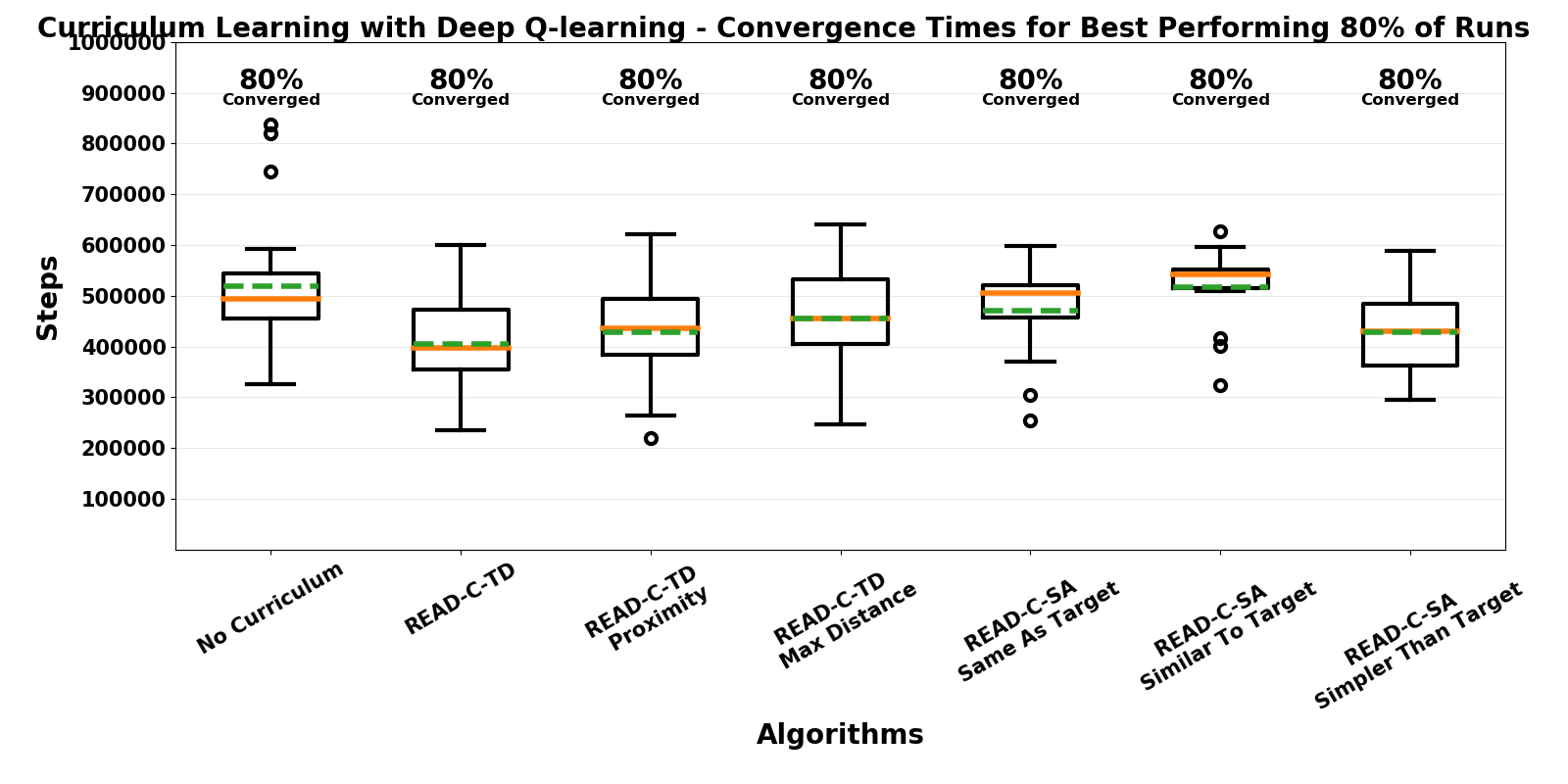}\label{fig:fig30}}
    \caption{ Box Plots for the Convergence Times of the Algorithms in Parking Domain.}
    \label{fig:fig31}
\end{figure*}

Figure~\ref{fig:fig25} contains the comparison of \readcsa{}, \readctd{} and advantage actor-critic algorithm \citep{a2c} for the parking domain specified in \citet{highway-env}, Figure~\ref{fig:fig26} shows the variants of \readctd{} and Figure~\ref{fig:fig27} displays different regressors used for \readcsa{}. Figure~\ref{fig:fig28} performs 25 runs of each algorithm similar to before. We do not present the results for different numbers of clusters as it does not add any additional significant findings to our results. Finally, we present the box plots showing the convergence times of each algorithm for a given cumulative reward value for 25 runs (Figure~\ref{fig:fig31}).

\readctd{} performs better than \readcsa{} and A2C in the parking environment while \readcsa{} still shows slight performance improvements over the baseline (Figure~\ref{fig:fig25}). The main reason for this difference between the parking and the grid world environments is that since the parking domain contains continuous state and action spaces, and \readc{} uses relative entropy for continuous variables, the error rate for the GBM regressor ends up being higher than before causing \readcsa{} to fall behind \readctd{} in certain cases. \readctd{} also performs better than its variants in this domain although \readctd{} + proximity almost gives the same performance results as \readctd{} (Figure~\ref{fig:fig26}). Most of the algorithms display overlapping confidence intervals but the confidence intervals for the baseline algorithm still remains significantly below the variants of \readc{} showing that \readc{} is able to provide improvement in learning efficiency for both continuous and discrete domains (Figure~\ref{fig:fig26}). \readcsa{} with source task simpler than the target task shows better performance than the other two competing regressors showing consistent results with the grid world domains (Figure~\ref{fig:fig27}). Although the gap between \readc{} and the baseline algorithm is narrower in these results, there still exists a benefit of using \readc{} in the continuous domain environments since most of \readc{} variants consistently manage to remain above the benchmark criteria aside from \readcsa{} with source task similar to the target task. 

For the convergence times, all \readc{} approaches show better mean and median performance than the A2C algorithm but they also produce worse convergence rates (Figure~\ref{fig:fig29}). The lower convergence rates in this case does not mean that the agent is not learning the task since we have seen in the line graphs that the asymptotic performance of the agent was better than the baseline algorithm but rather, the lower convergence rates indicate that the agent in some runs is not able to reach the reward value of -15 for 10 consecutive episodes despite improving its policy performance. However, if we look at the results obtained using the best performing 80\% runs, we see that all \readc{} algorithms, except READ-SA with source task similar to the target task, still produce better mean and median values while having the same convergence rate (Figure~\ref{fig:fig30}), meaning that the reason \readc{} shows low convergence rate in the first case is because of the 20\% poor performing runs. Overall, \readcsa{} manages to outperform its competition albeit it cannot guarantee convergence for all cases. \readctd{} shows better convergence rates than its variants and \readcsa{}, which is consistent with the results of the line graphs (Figure~\ref{fig:fig29}). Also, \readcsa{} with simpler source task than the target task obtains a better mean and median convergence rate than \readcsa{} with source task same as the target task and \readcsa{} with source task similar to the target task, which is consistent with the results we obtained in the Key-Lock domain (Figure~\ref{fig:fig30}). 

\section{Conclusion and Future Work}

We presented \readc{}, a novel approach to using relative entropy for automatic
curriculum generation. We detailed two versions of the approach, one using a
teacher to measure relative entropy and another self-calculating the relative
entropy without the teacher.  We adapted a curriculum generation criteria based on max policy change from the literature to
compare against \readc{} and evaluated performance on the key-lock domain, capture-the-flag domain and the parking domain using discrete and continuous action and state spaces.
Our results showed that \readc{} was able to outperform a proximity-aware randomly generated
curriculum, the comparison criteria, and learning the target task directly in
all our experiments. Further, \readcsa{} produced
better results than the baselines and \readctd{} in most cases even without having access to a teacher model.

\section*{Declarations}

\subsection*{Funding}
The authors did not receive support from any organization for the submitted work.
\subsection*{Conflict of interest/Competing interests}
The authors have no relevant financial or non-financial interests to disclose.
\subsection*{Ethics approval and consent to participate}
There are no ethical considerations which we feel must be specifically highlighted here. 
\subsection*{Consent for publication}
All authors consent to the publication of the materials presented in this paper.
\subsection*{Materials Availability}
Not applicable
\subsection*{Data and Code availability}
All relevant resources will be made available on Github upon publication. 
\subsection*{Author contribution}
Methodology: Muhammed Yusuf Satici; Formal analysis and investigation: Muhammed Yusuf Satici; Writing - original draft preparation: Muhammed Yusuf Satici; Writing - review and editing: Muhammed Yusuf Satici, Jianxun Wang, David Roberts; Supervision: David Roberts.

\bibliography{ref}
\end{document}